%% file: main.tex
\documentclass[10pt,twocolumn,letterpaper]{article}

\usepackage[pagenumbers]{cvpr} 

\usepackage{graphicx}
\usepackage{amsmath}
\usepackage{bbm}
\usepackage{amssymb}
\usepackage{booktabs}
\usepackage{multirow}
\usepackage{makecell}
\usepackage{bm}
\usepackage{pifont}
\usepackage{algorithm}
\usepackage{algpseudocode}
\usepackage[dvipsnames]{xcolor}

\usepackage[pagebackref,breaklinks,colorlinks]{hyperref}

\usepackage[capitalize]{cleveref}
\crefname{section}{Sec.}{Secs.}
\Crefname{section}{Section}{Sections}
\Crefname{table}{Table}{Tables}
\crefname{table}{Tab.}{Tabs.}

\newcommand{\beginsupp}{
        \setcounter{table}{0}
        \renewcommand{\thetable}{S\arabic{table}}
        \setcounter{figure}{0}
        \renewcommand{\thefigure}{S\arabic{figure}}
        \setcounter{section}{0}
        \renewcommand{\thesection}{\Alph{section}}
     }

\def\cvprPaperID{5021} 
\def\confName{CVPR}
\def\confYear{2022}

\newcommand{\norm}[1]{\left\lVert#1\right\rVert}

\begin{document}

\title{Class Re-Activation Maps for Weakly-Supervised Semantic Segmentation}

\author{
Zhaozheng Chen$^{1}$ \quad Tan Wang$^{2}$ \quad Xiongwei Wu$^{1}$ \\
Xian-Sheng Hua$^{3}$ \quad Hanwang Zhang$^{2}$ \quad Qianru Sun$^{1}$ \\
\\
\small  $^{1}$Singapore Management University \quad  $^{2}$Nanyang Technological University \quad $^{3}$Damo Academy, Alibaba Group \\
\small  {\texttt{zzchen.2019@phdcs.smu.edu.sg}} \quad {\texttt{\{tan317, hanwangzhang\}@ntu.edu.sg}}\\
\small  {\texttt{xiansheng.hxs@alibaba-inc.com}} \quad  {\texttt{\{xwwu,qianrusun\}@smu.edu.sg}}
}

\maketitle

\input{sections/0_abstract}
\input{sections/1_introduction}
\input{sections/2_related}
\input{sections/3_method}
\input{sections/4_experiment}
\input{sections/5_conclusion}

{\small
\bibliographystyle{ieee_fullname}
\bibliography{main}
}

\clearpage
\input{sections/6_supp}

\end{document}

%% file: sections/0_abstract.tex
\begin{abstract}

Extracting class activation maps (CAM) is arguably the most standard step of generating pseudo masks for weakly-supervised semantic segmentation (WSSS). Yet, we find that the crux of the unsatisfactory pseudo masks is the binary cross-entropy loss (BCE) widely used in CAM. Specifically, due to the sum-over-class pooling nature of BCE, each pixel in CAM may be responsive to multiple classes co-occurring in the same receptive field. As a result, given a class, its hot CAM pixels may wrongly invade the area belonging to other classes, or the non-hot ones may be actually a part of the class. To this end, we introduce an embarrassingly simple yet surprisingly effective method: Reactivating the converged CAM with BCE by using softmax cross-entropy loss (SCE), dubbed \textbf{ReCAM}. Given an image, we use CAM to extract the feature pixels of each single class, and use them with the class label to learn another fully-connected layer (after the backbone) with SCE. Once converged, we extract ReCAM in the same way as in CAM. Thanks to the contrastive nature of SCE, the pixel response is disentangled into different classes and hence less mask ambiguity is expected. The evaluation on both PASCAL VOC and MS~COCO shows that ReCAM not only generates high-quality masks, but also supports plug-and-play in any CAM variant with little overhead. Our code is public at \href{https://github.com/zhaozhengChen/ReCAM}{https://github.com/zhaozhengChen/ReCAM}.

\end{abstract}

%% file: sections/1_introduction.tex
\section{Introduction}
\label{sec_intro}

Weakly-supervised semantic segmentation (WSSS) aims to lower the high cost in annotating ``strong'' pixel-level masks by using ``weak'' labels instead, such as scribbles~\cite{scribble1,scribble2}, bounding boxes~\cite{bbox1,bbox2}, and image-level class labels~\cite{sec,irn,auxiliary,complementary,group,matters}. The last one is the most economic yet challenging budget and thus is our focus in this paper. A common pipeline has three steps: 1) training a multi-label classification model with the image-level class labels; 2) extracting the class activation map (CAM)~\cite{cam} of each class to generate a 0-1 mask, with potential refinement such as erosion and expansion~\cite{irn,anti}; and 3) taking all-class masks as pseudo labels to learn the segmentation model in a standard fully-supervised fashion~\cite{deeplabv2,deeplabv3+,upernet}. There are different factors affecting the performance of the final segmentation model, but the classification model in the first step is definitely the root. We often observe two common flaws. In the CAM of an object class A, there are 1) \underline{{false positive pixels}} that are activated for class A but have the actual label of class B, where B is usually a confusing class to A rather than \texttt{background}---a special class in semantic segmentation; and 2) \underline{{false negative pixels}} that belong to class A but are wrongly labeled as \texttt{background}.

\input{figures/figure_tisser}

\noindent
\textbf{Findings.}
We point out that these flaws are particularly obvious when the model is trained with the binary cross-entropy (BCE) loss with sigmoid activation function. Specifically, the sigmoid function is $\frac{e^{x}}{e^{x}+1}$ where $x$ denotes the prediction logit of any individual class. The output is fed into the BCE function to compute a loss. This loss represents the penalty strength for misclassification corresponding to $x$. The BCE loss is thus not class mutually exclusive---the misclassification of one class does not penalize the activation on others. This is indispensable for training multi-label classifiers. However, when extracting CAM via these classifiers, we see the drawbacks: non-exclusive activation across different classes (resulting in \underline{{false positive pixels}} in CAM); and the activation on total classes is limited (resulting in \underline{{false negative pixels}}) since partial activation is shared.

\noindent
\textbf{Motivation.}
We conduct a few toy experiments to empirically show the poor quality of CAM when using BCE. We pick single-label training images in MS COCO 2014~\cite{mscoco} (about 20\% in the \texttt{train} set) to train 5-class and 80-class classifiers, respectively, where for 5-class, we pick 5 hoofed animal classes (e.g., \texttt{horse} and \texttt{cow})
that suffer from the confused activation. We train every model using two losses, respectively: BCE loss and softmax cross-entropy (SCE) loss---the most common one for classification. We use the single-label images in \texttt{val} set to evaluate models' classification performance, as shown in Figure~\ref{fig:tisser}~(a), and use the single-label images in both \texttt{train} and \texttt{val} sets to inspect models' ability of activating correct regions on the objects---the quality of CAM, as compared in Figure~\ref{fig:tisser}~(b).

Intrigued, 1) for 80-class models, BCE and SCE yield equal-quality classifiers but clearly different CAMs, and 2) the CAMs of SCE models are of higher mIoU, and this superiority is almost maintained for validation images. A small yet key observation is that for 5 hoofed animal classes, BCE shows weaker to classify them.
We point out this is because the sigmoid activation function of BCE does not enforce class-exclusive learning, confusing the model between similar classes. However, SCE is different. Its softmax activation function $\frac{e^{x}}{e^{x}+\Sigma_y{e^{y}}}$, where $y$ denotes the prediction of any negative class, explicitly enforces class-exclusion by using exponential terms in the denominator. SCE encourages to improve the logit of ground truth and penalizes others simultaneously.
This makes two effects on CAM: 1) reducing \underline{{false positive pixels}} which confuse the model among different classes; and 2) encouraging the model to explore class-specific features that reduce \underline{{false negative pixels}}. We show the empirical evidence in Figure~\ref{fig:tisser}~(b) where the mIoU improvements by SCE over BCE are especially significant for 5-hoofed. \textbf{Please note} that the functions of BCE and SCE are different. 
To give more concrete comparison between them, we elaborate the comparison between their produced gradients in Section~\ref{sec_discussion_loss}, theoretically and empirically.

\noindent
\textbf{Our Solution.}
Our intuition is to use SCE loss function to train a model for CAM. 
However, directly replacing BCE with SCE does not make sense for multi-label classification tasks where the probabilities of different classes are not independent~\cite{multilabel,multilabel_reduc}.
Instead, we use SCE as an additional loss to \emph{Re}activate the model and generate \emph{Re}CAM.
Specifically, when the model converges with BCE, for every individual class labeled in the image, we extract the CAM in the format of normalized soft mask, i.e., without hard thresholding~\cite{cam, edam}.
We apply all masks on the feature (i.e., the feature map block output by the backbone), respectively, each ``highlighting''
the feature pixels contributing to the classification of a specific class. In this way, we branch the multi-label feature to a set of single-label features. We can thus use these features (and labels) to train a multi-class classifier with SCE, e.g., by plugging another fully-connected layer after the backbone. The SCE loss penalizes any misclassification caused by either poor features or poor masks. Then, backpropagating its gradients
improves both. Once converged, we extract ReCAM in the same way of CAM.

\noindent
\textbf{Empirical Evaluations.}
To evaluate the ReCAM, 
we conduct extensive WSSS experiments on two popular benchmarks of semantic segmentation, PASCAL VOC 2012~\cite{voc} and MS COCO 2014~\cite{mscoco}. A standard pipeline of WSSS is to use CAM~\cite{cam} as seeds and then deploy refinement methods such as AdvCAM~\cite{anti} or IRN~\cite{irn} to expand the seeds to pseudo masks---the labels used to train the segmentation model. We design the following comparisons to show the generality and superiority of ReCAM. 1)~\emph{ReCAM as seeds, too}. We extract ReCAM and use refinement methods afterwards, showing that the superiority over CAM is maintained after strong refinement steps. 2)~\emph{ReCAM as another refinement method.} We compare ReCAM with existing refinement methods, regarding the quality of generated masks as well as the computational overhead added to baseline CAM~\cite{cam}. In the stage of learning semantic segmentation models, we use the ResNet-based DeepLabV2~\cite{deeplabv2}, DeepLabV3+~\cite{deeplabv3+} and the transformer-based UperNet~\cite{upernet}.

\noindent
\textbf{Our Contributions} in this paper are thus two-fold. 
1) A simple yet effective method ReCAM for generating pseudo masks for WSSS.
2) Extensive evaluations of ReCAM on two popular WSSS benchmarks, with or without incorporating advanced refinement methods~\cite{irn,anti}.

%% file: figures/figure_tisser.tex
\begin{figure}[t]
    \centering
    \footnotesize
    \includegraphics[width=.98\linewidth]{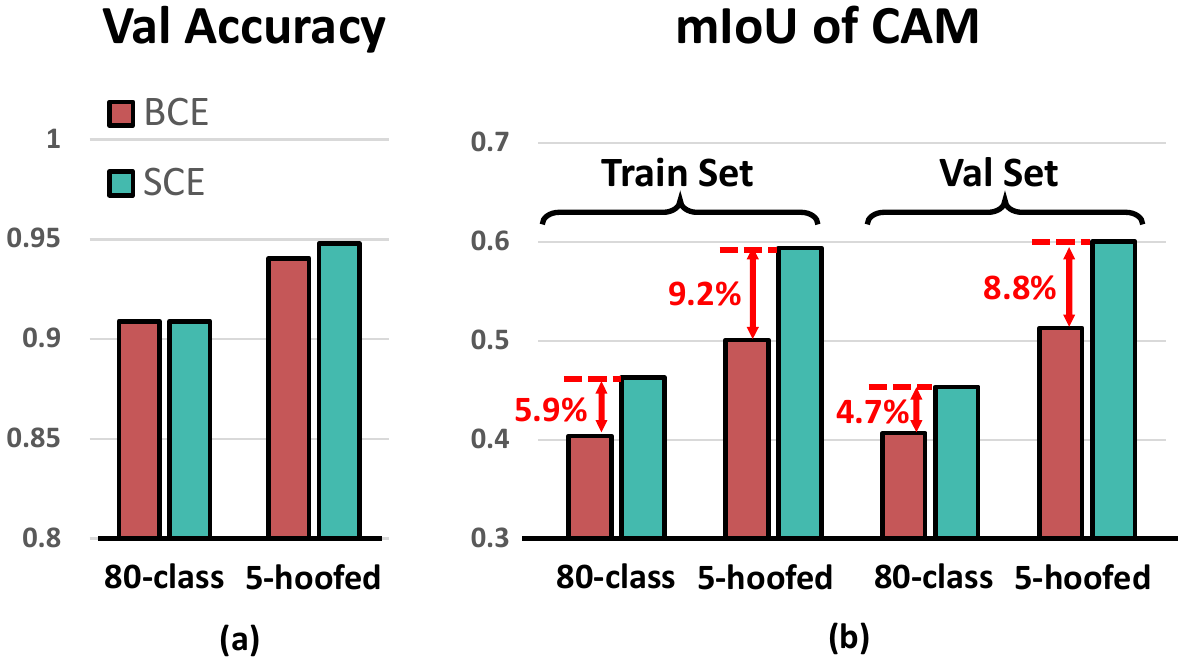}
    \vspace{-3mm}
    \caption{We train two models respectively using binary cross entropy (BCE) and softmax cross entropy (SCE) losses. Our \texttt{train} and \texttt{val} sets contain only single-label images of MS~COCO~\cite{mscoco}. ``80-class'' model uses the complete label set. ``5-hoofed'' model is trained on only the samples of 5 hoofed animals each causing false positive flaws to another, e.g., between \texttt{cow} and \texttt{horse}.}
    \vspace{-3mm}
    \label{fig:tisser}
\end{figure}

%% file: sections/2_related.tex
\section{Related Works}
\label{sec_related}

The training of multi-label classification and semantic segmentation models are almost uniform in the works of WSSS.
Below, we introduce only the variants for seed generation and mask refinement.

\noindent
\textbf{Seed Generation.}
Vanilla CAM~\cite{cam} first scales the feature maps (e.g., output by the last residual block) by using the FC weights learned for each individual class. Then, it produces the seed masks by channel-wise averaging, spatial-wise normalization, and hard thresholding (see Section~\ref{sec_preli}). Based on this CAM, there are improved methods. GAIN~\cite{gain} applies the CAM on original images to generate masked images, and minimizes the model prediction scores on them, forcing the model to capture the features in other regions (outside the current CAM) in the new training. A similar idea was used in erasing-based methods~\cite{adv_erasing,cse,selferasing,adv_erasing2}. The difference is erasing methods directly perturbed the regions (inside the CAM) and fed the perturbed images into the model to generate the next-round CAM that is expected to capture new regions. Score-CAM~\cite{scorecam} is a different CAM method. It replaces the FC weights used in vanilla CAM with a new set of scores predicted from the images masked by channel-wise (not class-specific) activation maps. EDAM~\cite{edam} is a recent work of using CAM-based perturbation to optimize an additional classifier.One may argue that our ReCAM is similar to EDAM. We highlight two differences. 1) EDAM uses an extra layer to produce class-specific soft masks, while our soft masks are simply from the byproducts of CAM without needing any parameters. 2) EDAM still uses BCE loss for training with perturbed input,
while we inspect the limitations of BCE and propose a different training method by leveraging SCE (see Section~\ref{sec_discussion_loss}).

\noindent
\textbf{Mask Generation.}
Seed masks generated by CAM or its variants can go through a refinement step. One category of refinement methods~\cite{psa,irn,auxiliary,bes} propagate the object regions in the seed to semantically similar pixels in the neighborhood. It is achieved by the random walk~\cite{randomwalk} on a transition matrix where each element is an affinity score. The related methods have different designs of this matrix. PSA~\cite{psa} is an AffinityNet to predict semantic affinities between adjacent pixels. IRN~\cite{irn} is an inter-pixel relation network to estimate class boundary maps based on which it computes affinities. Another method is BES~\cite{bes} that learns to predict boundary maps by using CAM as pseudo ground truth. All these methods introduced additional network modules to vanilla CAM. Another category of refinement methods~\cite{nsrom,dsrg,attenbn,ficklenet,splitting,suppression} utilize saliency maps~\cite{saliency1,saliency2}. EPS~\cite{eps} proposed a joint training strategy to combine CAM and saliency maps. EDAM~\cite{edam} introduced a post-processing method to integrate the confident areas in the saliency map into CAM. In experiments, we plug ReCAM in them to evaluate its performance with additional saliency data. A more recent category of methods leverage iterative post-processing to refine CAM. OOA~\cite{ooa} ensembles the CAM generated in multiple training iterations. CONTA~\cite{conta} iterated through the whole process of WSSS including a sequence of model training and inference. AdvCAM~\cite{anti} used the gradients with respect to the input image to perturb the image, and iteratively find newly activated pixels. Overall, these refinement methods are based on the seed generated by CAM~\cite{cam}. Our ReCAM is a method of leveraging SCE to  reactivate more pixels in CAM, and is thus convenient to incorporate it. We conduct extensive plug-and-play experiments in Section~\ref{sec_exper}.

Other ideas of improving CAM include ICD~\cite{icd} that learned intra-class boundaries on feature manifolds,
SC-CAM~\cite{sub_category} that learned fine-grained classification models (with pseudo fine-grained labels);
and SEAM~\cite{seam} that enforced the consistency of CAM extracted from different transformations of the image.
A recent work RIB~\cite{rib} did a careful analysis based on the theory of information bottleneck, and proposed
to retrain the multi-label classification model without the last activation function. 
Our ReCAM does not remove any activation function but adds a softmax activation based loss (SCE), as shown in Figure~\ref{fig:framework_recam}.
Another difference is in the inference stage. RIB needs $10$ iterations of feeding forward and backward for each test image, but ReCAM feeds forward the image only once. 
For example, on PASCAL VOC 2012~\cite{voc} dataset, RIB costs $8$ hrs in its inference (its training cost is the same as vanilla CAM), while our total cost over vanilla CAM is only $0.6$ hrs.

%% file: sections/3_method.tex
\section{Preliminaries}
\label{sec_preli}

\noindent
\textbf{CAM.}
The first step of CAM~\cite{cam} is to train a multi-label classification model with global average pooling (GAP) followed by a prediction layer (e.g., the FC layer of a ResNet~\cite{resnet}). The prediction loss on each training example is computed by BCE function in the following formula:
{\small
\begin{equation}
    \mathcal{L}_{bce}=-\frac{1}{K} \sum_{k=1}^{K} y \left[ k \right] \log \sigma\left(z[k]\right)+\left(1-y[k]\right) \log \left[1-\sigma\left(z[k]\right)\right],
\label{eq:bce}
\end{equation}}

\noindent
where \(z[k]\) denotes the prediction logit of the \(k\)-th class, \(\sigma(\cdot)\) is the sigmoid function, and \(K\) is the total number of foreground object classes (in the dataset). \(y[k]\in\{0,1\}\) is the image-level label for the \(k\)-th class, where 1 denotes the class is present in the image and 0 otherwise.

Once the model converges, we feed the image $\bm{x}$ into it to extract the CAM of class $k$ appearing in $\bm{x}$:
\begin{equation} \label{equation:cam}
    \operatorname{CAM}_k(\bm{x})=\frac{\operatorname{ReLU}\left(\bm{A}_k\right)}{\max \left(\operatorname{ReLU}\left(\bm{A}_k\right)\right)}, \bm{A}_k=\mathbf{w}_{k}^{\top}f(\bm{x}),
\end{equation}
where \(\mathbf{w}_{k}\) denotes the classification weights (e.g., the FC layer of a ResNet) corresponding to the \(k\)-th class, and \(f(\bm{x})\) represents the feature maps of \(\bm{x}\) before the GAP.

\textbf{Please note} that for simplicity, we assume the classification head of the model is always a single FC layer, and use \(\mathbf{w}\) to denote its weights in the following.

\noindent
\textbf{Pseudo Masks.} 
There are a few options to generate pseudo masks from CAM: 
1) thresholding CAM to be 0-1 masks; 
2) refining CAM with IRN~\cite{irn}---a widely used refinement method;
3) iteratively refining CAM through the classification model, e.g., using AdvCAM~\cite{anti};
and 4) cascading options 3 and 2. 
In Figure~\ref{fig:framework_two_steps}, we illustrate these options with our ReCAM plugged in. We elaborate these in Section~\ref{sec_recam_pipeline}.

\noindent
\textbf{Semantic Segmentation.} 
This is the last step of WSSS. We use the pseudo masks to train the semantic segmentation model in a fully-supervised way.
The objective function is as follows:
{\small
\begin{equation}
    \mathcal{L}_{ss}=-\frac{1}{HW} \sum_{i=1}^{H} \sum_{j=1}^{W} \frac{1}{K\!+\!1} \sum_{k=1}^{K+1} y_{i,j}[k]  \log \frac{\exp (z_{i,j}[k])}{\sum_{k} \exp (z_{i,j}[k])},
\end{equation}}

\noindent
where $y_{i,j}$ and $z_{i,j}$ denote the label and the prediction logit at pixel $(i,j)$, respectively. $y_{i,j}[k]$ and $z_{i,j}[k]$ denote the $k$-th element of $y_{i,j}$ and $z_{i,j}$, respectively. $H$ and $W$ are the height and width of the image. $K$ is the total number of classes. $K\!+\!1$ means including the \texttt{background} class.

For implementation, we deploy DeepLab variants~\cite{deeplabv2, deeplabv3+} with ResNet-101~\cite{resnet}, following related works~\cite{irn,conta,anti,rib}. In addition, we employ a recent model UperNet~\cite{upernet} with a stronger backbone---Swin Transformer\footnote{Please note that we did not implement our method on transformer-based classification models and will take this as the future work.}~\cite{swin}.

\input{sections/3_method_part2}

%% file: sections/3_method_part2.tex
\section{Class Re-Activation Maps (ReCAM)}
\label{sec_recam}

\input{figures/framework_two_steps}

\input{figures/framework_recam}

In Section~\ref{sec_recam_pipeline}, we elaborate our method of reactivating the classification model and extracting ReCAM from it.
Note that we also use ``ReCAM'' to name our method. In Section~\ref{sec_discussion_loss}, we justify the advantages of class-exclusive learning in ReCAM, by comparing the gradients of SCE with BCE theoretically and empirically.

\subsection{ReCAM Pipeline}
\label{sec_recam_pipeline}

\noindent
\textbf{Backbone and Multi-Label Features.} 
We use a standard ResNet-50~\cite{resnet} as our backbone (i.e., feature encoder) to extract features, following related works~\cite{irn,anti,rib,conta}.

Given an input image $\bm{x}$ and its multi-hot class label $\bm{y}\in\{0,1\}^{K}$, we denote the output of feature encoder as $f(\bm{x}) \in \mathbb{R}^{W \times H \times C}$. $C$ denotes the number of channels, $H$ and $W$ denote the height and width, respectively. $K$ is the total number of foreground classes in the dataset. Please note that in Figure~\ref{fig:framework_recam}, 1) the feature extraction process is omitted for conciseness; and 2) the feature $f(\bm{x})$ is written as $\bm{f}$ in the upper block and usually represents multiple objects.

\noindent
\textbf{FC Layer-1 with BCE Loss.} 
In the conventional model of CAM, the feature $f(\bm{x})$ first goes through a GAP layer and the result is fed into a FC layer to make prediction~\cite{cam}. Hence, the prediction logits can be denoted as
\begin{equation}
    \bm{z} = \operatorname{FC_1}(\operatorname{GAP}(f(\bm{x}))).
\end{equation}
Then, $\bm{z}$ and the image-level labels $\bm{y}$ are used to compute a BCE loss. An element-wise formula is given in Eq.~\eqref{eq:bce}.

\noindent
\textbf{Extracting CAM.}
We extract the CAM for each individual class $k$ given the feature $f(\bm{x})$ and the corresponding weights $\mathbf{w}_k$ of the FC layer, as formulated in Eq.~\eqref{equation:cam}. For brevity, we denote the $\operatorname{CAM}_k(\bm{x})$ as $\bm{M}_k \in \mathbb{R}^{W \times H}$. 

\noindent
\textbf{Single-Label Feature.}
As shown in Figure~\ref{fig:framework_recam}, we use $\bm{M}_k$ as a soft mask to apply on $f(\bm{x})$ to extract the class-specific feature $f_k(\bm{x})$. We compute the element-wise multiplication between $\bm{M}_k$ and each channel of $f(\bm{x})$ as follows,
\begin{equation}
\label{eq:masked_feature}
    f_k^c(\bm{x}) = \bm{M}_k \otimes f^c(\bm{x}),
\end{equation}
where $f^c(\bm{x})$ and $f_k^c(\bm{x})$ indicate the single channel before and after the multiplication (by using $\bm{M}_k$), $c$ ranges from $1$ to $C$ and $C$ is number of feature maps (i.e., channels). The feature map block $f_k(\bm{x})$ (each contains $C$ channels) corresponds to the examples $\bm{f}_1, \bm{f}_2, \bm{f}_{3}$ in Figure~\ref{fig:framework_recam}.

\noindent
\textbf{FC Layer-2 with SCE Loss.} 
Each $f_k(\bm{x})$ has a single object label (i.e., a one-hot label where the $k$-th position is 1). Then, we feed it to FC Layer-2 (see Figure~\ref{fig:framework_recam}) to learn multi-class classifier, so we have new prediction logits for $\bm{x}$ as:
\begin{equation}
    \bm{z}_{k}' = \operatorname{FC_2}(\operatorname{GAP}(f_k(\bm{x}))),
\end{equation}
where $\operatorname{FC_2}$ has the same architecture as $\operatorname{FC_1}$.

By this way, we succeed to convert the BCE-based model on multi-label images to the SCE-based model on single-label features. The SCE loss is formulated as:
\begin{equation}
    \mathcal{L}_{sce}=-\frac{1}{\sum_{i=1}^{K}\bm{y}[i]} \sum_{k=1}^{K} \bm{y}[k] \log \frac{\exp (\bm{z}_{k}'[k])}{\sum_{j} \exp (\bm{z}_{k}'[j])},
    \label{eq:sce_loss}
\end{equation}
where $\bm{y}[k]$ and $\bm{z}_{k}'[k]$ denotes the $k$-th elements of $\bm{y}$ and $\bm{z}_{k}'$, respectively. 
We use the gradients of $\mathcal{L}_{sce}$ to update the model including the backbone.

Therefore, our overall objective function for reactivating the BCE model is as follows:
\begin{equation}
    \label{eq:loss_ours}
    \mathcal{L}_{ReCAM}= \mathcal{L}_{bce} + \lambda \mathcal{L}_{sce},
\end{equation}
where $\lambda$ is to balance between BCE and SCE. Please note the re-optimization of FC1 with $\mathcal{L}_{bce}$ is also included because we need to use FC1 to produce updated soft masks $\bm{M}_k$ during the learning.

\noindent
\textbf{Extracting ReCAM.} 
After the reactivation, we feed the image $\bm{x}$ into it to extract its ReCAM of each class $k$ as follows,
\begin{equation}
 \operatorname{ReCAM}_k(\bm{x})=\frac{\operatorname{ReLU}\left(\bm{A}_k\right)}{\max \left(\operatorname{ReLU}\left(\bm{A}_k\right)\right)}, \bm{A}_k={\mathbf{w}_{k}''}^{\top}f(\bm{x}),
\label{eq:recam}
\end{equation}
where $\mathbf{w}_{k}''$ denotes the classification weights corresponding to the \(k\)-th class. As we have two FC layers, our implementation takes $\mathbf{w}''$ optional as: 1) $\mathbf{w}$, 2) $\mathbf{w}'$, 3)$\mathbf{w} \oplus \mathbf{w}'$, or 4)$\mathbf{w} \otimes \mathbf{w}'$, where $\oplus$ and $\otimes$ are element-wise addition and multiplication, respectively. 
We show the performances of these options in Section~\ref{sec_result}.

\noindent\textbf{Refining ReCAM (Optional).}
As introduced in Section~\ref{sec_preli}, there are a few options to refine ReCAM:
\textbf{1) AdvCAM}~\cite{anti} iteratively refines ReCAM by perturbing images $\bm{x}$ through adversarial climbing:
{\small
\begin{equation}
\begin{aligned}
\bm{x}^{t} &= \bm{x}^{t-1}+\xi \nabla_{\bm{x}^{t-1}} \mathcal{L}_{adv}, \\
\mathcal{L}_{adv} &= \bm{y}^{t-1}[k]-\sum_{j \in K \backslash k} \bm{y}^{t-1}[j] \\
&-\mu \left\|\mathcal{M} \otimes\left|\operatorname{ReCAM}_{k}\left(\bm{x}^{t-1}\right)-\operatorname{ReCAM}_{k}\left(\bm{x}^{0}\right)\right|\right\|_{1},
\label{eq:advcam-recam}
\end{aligned}
\end{equation}}

\noindent
where $t\in [1,T]$ is the adversarial step index, $\bm{x}^t$ is the manipulated image at the $t$-th step. $k$ and $j$ are the positive and negative classes, respectively. $\xi$ and $\mu$ are hyper-parameters (same as in~\cite{anti}).
$\mathcal{M}=\mathbbm{1}\left(\operatorname{ReCAM_k}\left(\bm{x}^{t-1}\right)>0.5\right)$ is a restricting mask of ReCAM for regularization. The final refined activation map $\bm{M}'_k=\frac{\sum_{t=0}^{T} \operatorname{ReCAM}_k\left(\bm{x}^{t}\right)}{\max \sum_{t=0}^{T} \operatorname{ReCAM}_k\left(\bm{x}^{t}\right)}$, note that here we follow AdvCAM~\cite{anti} to use ReCAM without max normalization in Eq.~\eqref{eq:recam}.
\textbf{2) IRN}~\cite{irn} takes ReCAM as the input and trains an inter-pixel relation network (IRNet) to estimate the class boundary maps $\mathcal{B}$. Here, we omit the training details of IRNet for brevity. Then, it applies a random walk to refine ReCAM with $\mathcal{B}$ and the transition probability matrix $\mathbf{T}$:
\begin{equation}
    \label{eq:irn}
    \operatorname{vec}\left(\bm{M}'_k\right)=\mathbf{T}^{t} \cdot \operatorname{vec}\left(\operatorname{ReCAM_k}(\bm{x}) \otimes(1-\mathcal{B})\right),
\end{equation}
where $t$ denotes the number of iterations and $\operatorname{vec}(\cdot)$ represents vectorization.
Finally, we use $\{\bm{M}'_k\}$ 
as pixel-level labels of the image, where $k$ denotes every positive class in the image, to train semantic segmentation models.

\subsection{Justification: BCE \emph{vs} CE}
\label{sec_discussion_loss}

In this section, we justify the advantages of introducing SCE loss in ReCAM. We compare the effects of SCE and BCE on optimizing the classification model, theoretically and empirically.

For any input image, let $\bm{z}$ denote the prediction logits and $\bm{y}$ as the one-hot label. 
Based on the derivation chain rule, the gradients of BCE and SCE\footnote{The SCE loss here is the vanilla SCE rather than Eq.~\eqref{eq:sce_loss}.} losses on logits can be derived as:
\begin{equation}
\begin{aligned}
    & \nabla_{\bm{z}} \mathcal{L}_{bce} = \frac{\sigma_{sig}(\bm{z})-\bm{y}}{K}, \\
    & \nabla_{\bm{z}} \mathcal{L}_{sce} = \sigma_{sof}(\bm{z})-\bm{y},
\label{eq:all_grad}
\end{aligned}
\end{equation}
where $\sigma_{sig}$ and $\sigma_{sof}$ represent sigmoid and softmax functions, respectively.

\noindent\textbf{Theoretically.}
For the ease of analysis, we consider the binary-class ($K=2$) situation with the positive class $p$ and negative class $q$. 
Eq.~\eqref{eq:all_grad} can be further derived as:
{\small
\begin{equation}
\begin{aligned}
    &\text{\ding{172}}~~  \nabla_{z_p} \mathcal{L}_{bce} = \frac{-1}{2+2e^{z_{p}}}  ~~~~~~
    \text{\ding{173}}~~ \nabla_{z_q} \mathcal{L}_{bce} = \frac{1}{2+2e^{-z_{q}}}  \\
    &\text{\ding{174}}~~  \nabla_{z_p} \mathcal{L}_{sce} = \frac{-1}{1+e^{z_{p}-z_{q}}} ~~~~~~
    \text{\ding{175}}~~ \nabla_{z_q} \mathcal{L}_{sce} = \frac{1}{1+e^{z_{p}-z_{q}}}
\label{eq:exp_grad}
\end{aligned}
\end{equation}}

\noindent
Then, we consider different situations of $z_p$ and $z_q$ to compare the \emph{magnitude} of gradient terms for both positive class $p$ (\ding{172} and \ding{174}) and negative class $q$ (\ding{173} and \ding{175}). 
a) $z_p \ll z_q$: the negative class logit is much larger than that of positive class. 
This case is quite rare and most are due to the false labeling. 
In this case, $\norm{\text{\ding{172}}}$ and $\norm{\text{\ding{173}}}$ are less than $0.5$, but $\norm{\text{\ding{174}}}$ and $\norm{\text{\ding{175}}}$ approach $1$---SCE converges faster.
b) $z_p \gg z_q$. This appears when model is converging. All the four gradient terms are close to $0$, which cannot tell any difference.

\input{figures/figure_grad}

Next, we consider the last and most confusing case: c) $z_p \approx z_q$. We split it into two subcases: c1) both $z_p$ and $z_q$ are large, e.g., around $10$ (as we observed in the MS~COCO ``5 hoofed'' experiments). %
We can find that the magnitude of the SCE loss gradients (i.e., $\norm{\text{\ding{174}}}$ and $\norm{\text{\ding{175}}}$) both approach $0.5$, while $\norm{\text{\ding{172}}}\approx 0$ and $\norm{\text{\ding{173}}}\approx 0.5$.
c2) $z_p$ and $z_q$ are small, e.g., around $-10$. $\norm{\text{\ding{174}}}$ and $\norm{\text{\ding{175}}}$ keep the same (as $0.5$), yet $\norm{\text{\ding{172}}}\approx 0.5$ and $\norm{\text{\ding{173}}}\approx 0$. We can find that, in both confusing cases, SCE loss yields gradients to encourage the prediction of positive class as well as to penalize the prediction of negative class.
The reason is the exponential terms in the denominator of the softmax function explicitly involve both classes. Based on this, SCE guarantees a class-exclusion learning---simultaneously improve the positive and suppress the negative, when confronting confusion. By contrast, in BCE each case focuses on either positive or negative class. It does not guarantee no reduction on the positive when penalizing the negative, or no promotion on the negative when encouraging the positive, leading to inefficient learning especially for confusing classes.

\noindent\textbf{Empirically.}
One may argue that larger magnitude of gradients might not directly lead to stronger optimization,
because common optimizers (e.g., Adam~\cite{adam}) use adaptive learning rates. To justify the effectiveness of SCE in practice, we monitor the gradients when running real models. In specific, we review the toy experiments of ``5 hoofed animals'' where the models are trained with the Adam optimizer. We compute the gradients of both BCE and SCE losses (produced via two independent models) with respect to each prediction logit. As shown in Figure~\ref{fig:figure_grad}, we show the gradients with respect to the logits of the target class (i.e., the only positive class $p$) and the confusing class (i.e., the negative class $q$ with the highest logit value), respectively. We can see that the gradients of SCE loss change more rapidly for both positive and negative classes, indicating that its model learns more actively and efficiently.

%% file: figures/framework_two_steps.tex
\begin{figure}[t]
    \centering
    \footnotesize
    \includegraphics[width=.98\linewidth]{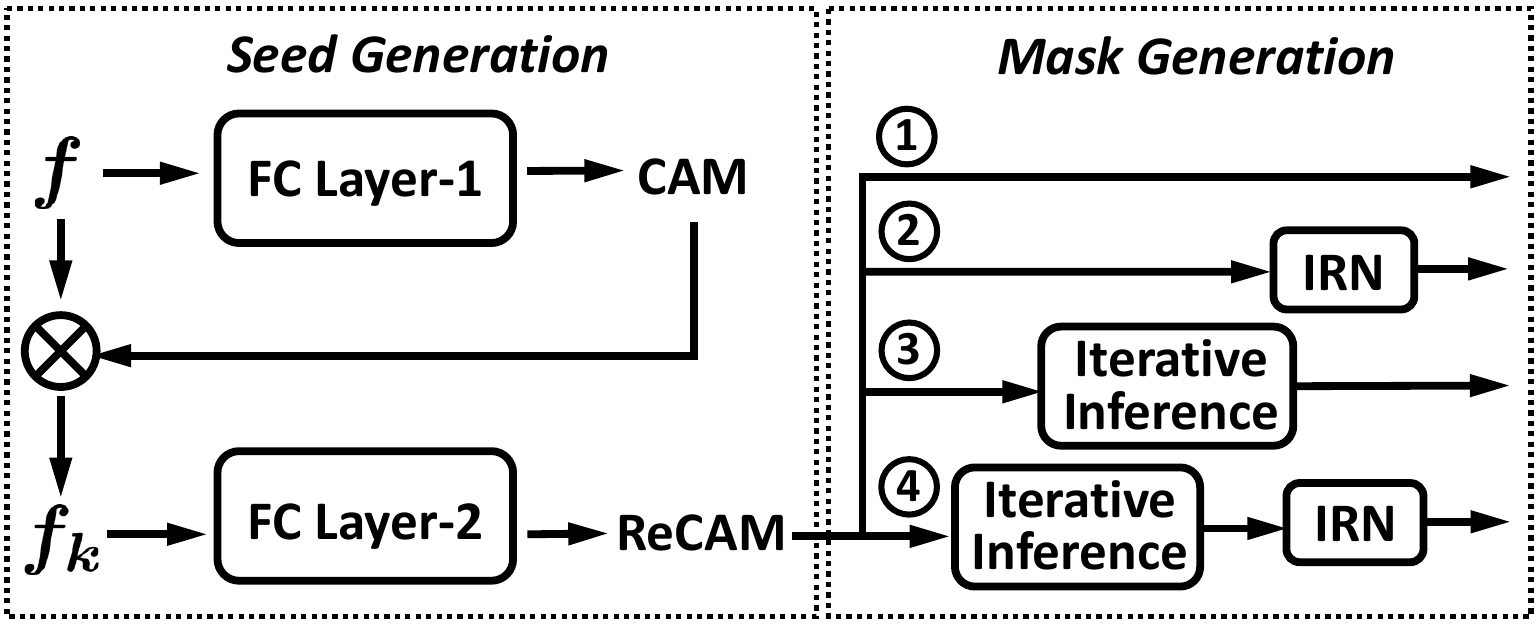}
    \vspace{-2mm}
    \caption{The pipeline of using ReCAM to generate pseudo masks for WSSS. 
    There are two steps, seed generation and mask generation, and our ReCAM is taken as a module plugged in seed generation step.
    The mask generation has a few options: 1) take the ReCAM directly as pseudo mask; 2) refine the ReCAM with the most common refinement method IRN~\cite{irn}; 
    3) iteratively infer better masks via the model of ReCAM; and 4) cascade options 3 and 2. The details of learning ReCAM model are shown in Figure~\ref{fig:framework_recam}. Table~\ref{table_plugin} shows the overall comparison results for these options.
    }
    \label{fig:framework_two_steps}
    \vspace{-4mm}
\end{figure}

%% file: figures/framework_recam.tex
\begin{figure}[t]
    \centering
    \footnotesize
    \includegraphics[width=.98\linewidth]{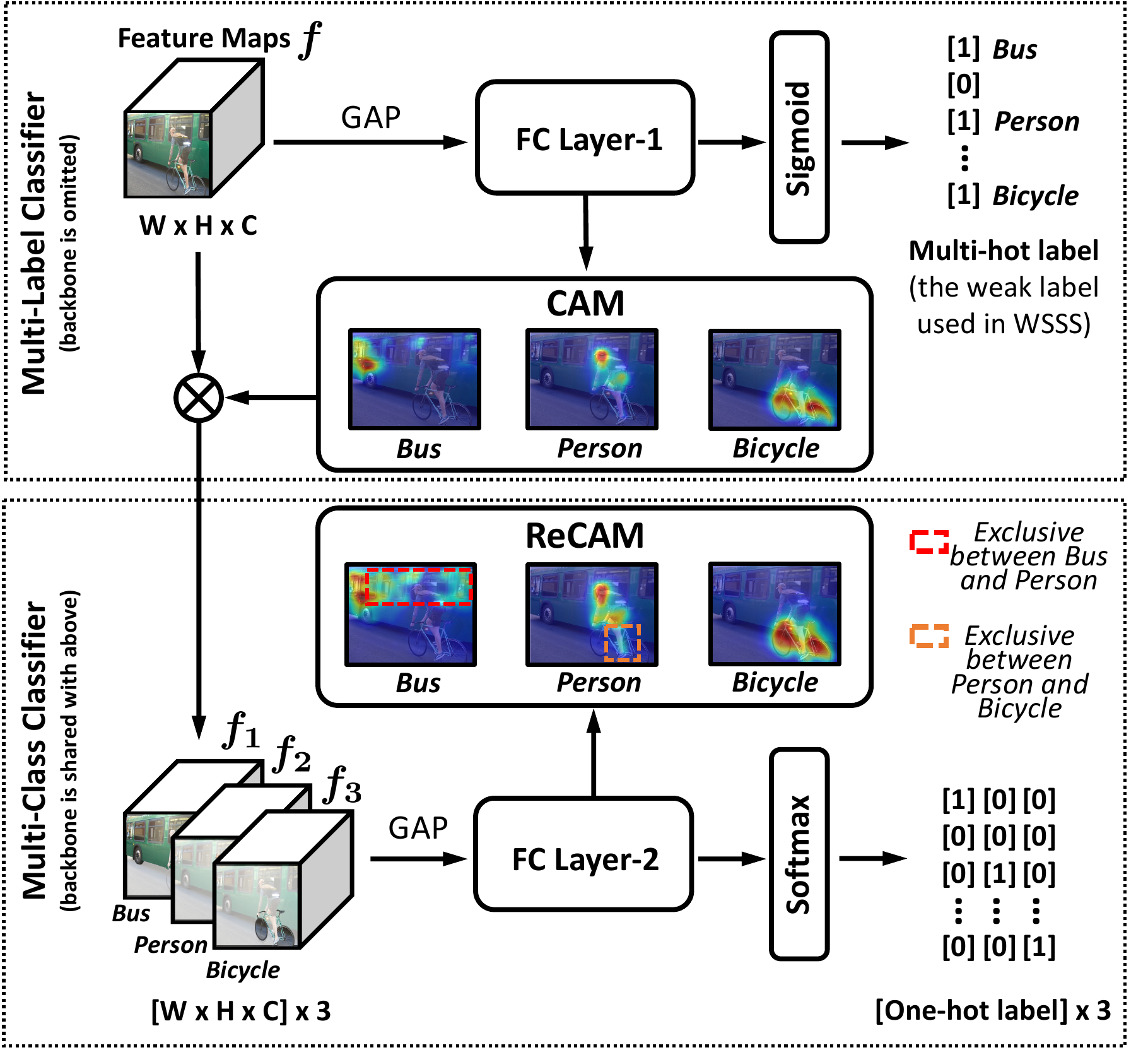}
    \vspace{-2mm}
    \caption{The training framework of ReCAM. In the upper block, it is the conventional training of multi-label classifiers using BCE. The feature extraction via backbone is omitted for conciseness. We extract the CAM for each class and then apply it (as a normalized soft mask) on the feature maps $\bm{f}$ to obtain the class-specific feature $\bm{f_k}$. In the lower block, we use $\bm{f_k}$ and its single label to learn multi-class classifiers with SCE loss. The gradients of this loss are backpropagated through the whole network including backbone.
    }
    \vspace{-4mm}
    \label{fig:framework_recam}
\end{figure}

%% file: figures/figure_grad.tex
\begin{figure}[t]
    \centering
    \footnotesize
    \includegraphics[width=.98\linewidth]{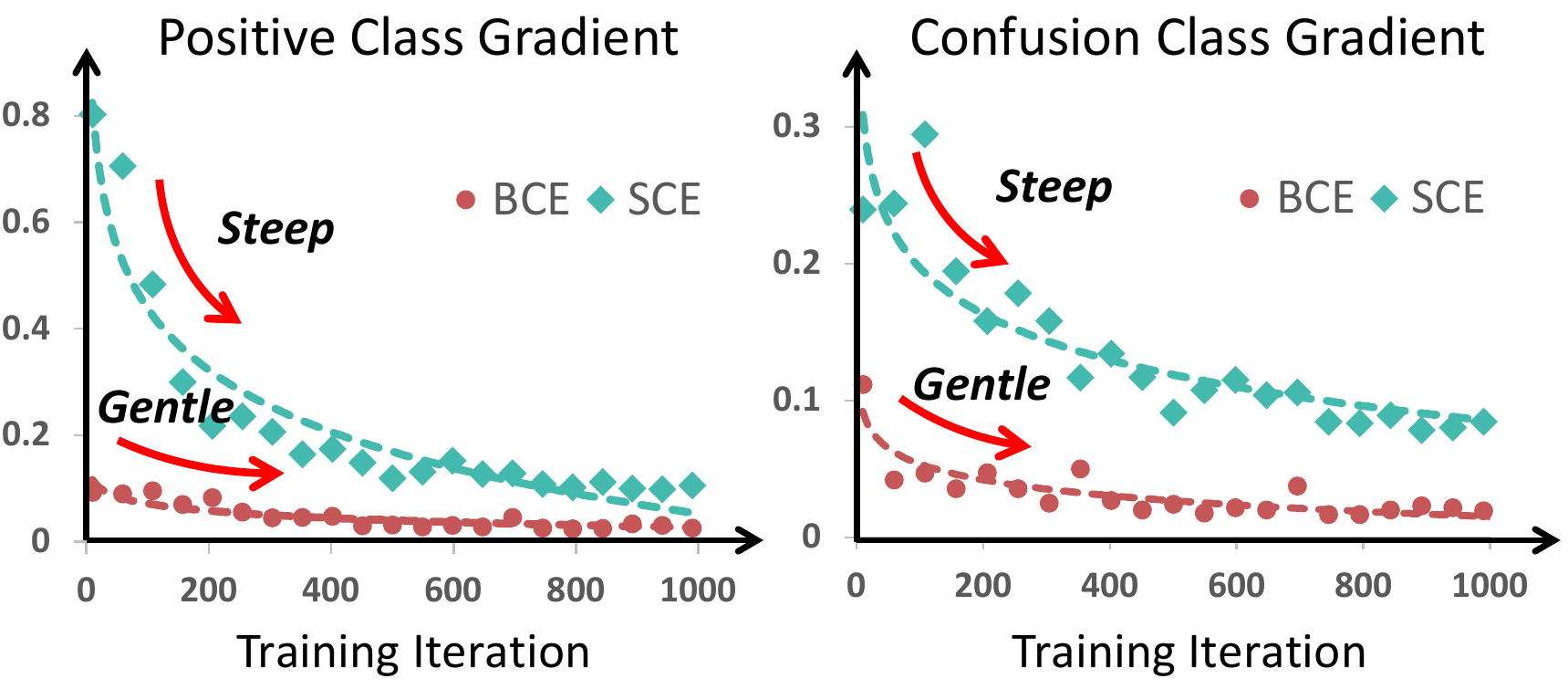}
    \vspace{-3mm}
    \caption{The gradients with respect to the logits of the target class (i.e., the only positive class $p$) and the confusing class (i.e., the negative class $q$ with the highest logit value).
    Both BCE and SCE models are trained with the 5 hoofed animal classes in MS~COCO \texttt{train} set.
    These gradients are calculated on the \texttt{val} set.
    }
    \vspace{-4mm}
    \label{fig:figure_grad}
\end{figure}

%% file: sections/4_experiment.tex
\section{Experiments}
\label{sec_exper}

\subsection{Datasets and Settings}

\noindent
\textbf{Datasets} include the commonly used PASCAL VOC 2012~\cite{voc} and MS~COCO 2014~\cite{mscoco}. VOC contains $20$ foreground object classes and $1$ \texttt{background} class. It has $1,464$, $1,449$, and $1,456$ samples in \texttt{train}, \texttt{val}, and \texttt{test} sets, respectively. Following related works~\cite{irn,conta,anti}, we used the enlarged training set with $10,582$ training images provided by Hariharen et al.~\cite{voc_aug}. MS~COCO contains $80$ object classes and $1$ \texttt{background} class. It has $80k$ and $40k$ samples, respectively, in its \textit{train} and \textit{val} sets. On both datasets, we used their image-level labels only during the training---the most challenging setting in WSSS.

\noindent
\textbf{Evaluation Metrics.} 
We have mainly two evaluation steps. \textit{Mask Generation.} We generate pseudo masks for the images in the \texttt{train} set and use their corresponding ground truth masks to compute the mIoU. \textit{Semantic Segmentation.} We train the segmentation model, use it to predict masks for the images in \texttt{val} or \texttt{test} sets, and compute the mIoU based on their ground truth masks. We also provide the results of F1 and pixel accuracy in the supplementary.

\noindent
\textbf{Network Architectures.}
For mask generation, we follow \cite{irn,conta,anti} to use ResNet-50 as backbone and its produced feature map size is $32\times 32\times 2048$. For semantic segmentation, we employ ResNet-101 (following \cite{irn,conta,anti}) and Swin Transformer~\cite{swin} (the first time in WSSS). Both are pre-trained on ImageNet~\cite{imagenet}. We incorporated ResNet-101 into DeepLabV2~\cite{deeplabv2} and DeepLabV3+~\cite{deeplabv3+}, where the results for the latter is in the supplementary due to space limits.
We incorporated Swin into UperNet~\cite{upernet}.

\noindent
\textbf{Implementation Details.}
For mask generation, we train FC Layer-1 with the same setting as in~\cite{irn}. We train FC layer-2 by: setting $\lambda$ as $1$ and $0.1$ on VOC and MS~COCO, respectively; running 4 epochs with the initial learning rate $5e^{-4}$ and polynomial learning rate decay on both datasets. We follow IRN~\cite{irn} to apply the same data augmentation and weight decay strategies. All hyper-parameters in Eq.~\eqref{eq:advcam-recam} and Eq.~\eqref{eq:irn} follow original AdvCAM~\cite{anti} and IRN~\cite{irn} paper. For the DeepLabV2 in the step of semantic segmentation, we use the same training settings as in~\cite{irn,anti,rib}. Please refer to the details in the supplementary. For the UperNet, the input image was first resized uniformly as $2,048\times 512$ with a ratio range from $0.5$ to $2.0$, and then cropped to be $512\times 512$ randomly before fed into the model. Data augmentation included horizontal flipping and color jitter. We trained the models for $40k$ and $80k$ iterations on VOC and MS~COCO datasets, respectively, with a common batch size of 16. We deployed AdamW~\cite{adamw} solver with an initial learning rate $6e^{-5}$ and weight decay as 0.01. The learning rate is decayed by a power of 1.0 according to the polynomial decay schedule.

\input{tables/table_ablation}
\input{tables/table_plugin}
\input{tables/table_seg}

\input{figures/figure_vis}

\subsection{Results and Analyses}
\label{sec_result}

\noindent
\textbf{SCE on FC Layer-1 (FC1) or Layer-2 (FC2).}
One may argue that SCE is not necessary to be applied on an additional classifier FC2. We conduct the experiments of using SCE on FC1 (i.e., \emph{w/o} FC2) and show the results in upper block of Table~\ref{table_ablation}. ``$\mathcal{L}_{bce}$ only'' is the baseline of using only BCE loss for FC1. ``$\mathcal{L}_{sce}$ only'' is to use SCE only for FC1, with modifying the original multi-hot labels to be normalized (summed up to $1$). For example, $[1,1,0,1,0]$ is modified as $[1/3,1/3,0,1/3,0]$. ``$\mathcal{L}_{sce}$ for single only'' is to apply BCE for learning multi-label images but SCE for single-label images (i.e., the subset of training images containing one object class). It shows that ``$\mathcal{L}_{sce}$ only'' performs the worst. This is because SCE does not make sense for multi-label classification tasks where the probabilities of different classes are not independent~\cite{multilabel}. ``$\mathcal{L}_{sce}$ for single only'' combines two losses to handle different images, which increases the complexity of the method. Moreover, it does not gain much, especially for MS~COCO dataset where there are a smaller number of single-label images and is a more general segmentation scenario in practice.

\noindent
\textbf{Using the Weights of FC1 and FC2 in Eq.\eqref{eq:recam}.} 
As we have two FC layers, our implementation of $\mathbf{w}''$ has a few options: 1) $\mathbf{w}$, 2) $\mathbf{w}'$, 3)$\mathbf{w} \oplus \mathbf{w}'$, or 4)$\mathbf{w} \otimes \mathbf{w}'$, where $\oplus$ and $\otimes$ are element-wise addition and multiplication, respectively. We show the results in the lower block of Table~\ref{table_ablation}. We can see that all options get better results than the baseline (i.e., ``$\mathcal{L}_{bce}$ only'' without FC2). ReCAM with $\mathbf{w} \otimes \mathbf{w}'$ achieves best performance on VOC. 
The reason is that the element-wise multiplication strengthens the representative feature maps and suppresses the confusing ones. Intriguingly, on MS~COCO, ReCAM with $\mathbf{w}$ achieves a better performance than $\mathbf{w} \otimes \mathbf{w}'$.  This is perhaps because the feature $f_k(\bm{x})$ input to FC2 is poor in this difficult dataset and FC2 is not well-trained\footnote{The \textcolor{red}{limitation} of ReCAM is its FC2 may overfit to the noisy features extracted by the poor backbone. We hope to tackle this in the future by leveraging strong pre-training methods that can upgrade the backbone.}. Based on these results, we use $\mathbf{w} \otimes \mathbf{w}'$ for all experiments on VOC, and $\mathbf{w}$ for MS~COCO.

It is worth highlighting that the effectiveness of ReCAM is validated on both datasets, if comparing any row in the second block with the first row in the Table~\ref{table_ablation}---any option of using ReCAM yields better masks than the baseline.

\input{figures/figure_lambda}

\noindent
\textbf{Effects of Different $\lambda$ Values.}
$\lambda$ in Eq.~\eqref{eq:loss_ours} controls the balancing between BCE and SCE. We study the pseudo mask quality (mIoU) of ReCAM by traversing the value of $\lambda$ on VOC, as shown in Figure~\ref{fig:lambda}~(a). We can observe that the optimal value of $\lambda$ is $1$, but the difference is not significant when using other values, i.e., ReCAM is not sensitive to $\lambda$.
Please kindly refer to supplementary materials for more sensitivity analysis, e.g., on learning rates.

\input{sections/4_experiment_part2}

%% file: tables/table_ablation.tex
\setlength{\tabcolsep}{2.2mm}{
\renewcommand\arraystretch{1.1}
\begin{table}
  \centering
    \scalebox{0.9}{
  \begin{tabular}{llcc}
    \toprule
        \multicolumn{2}{c}{Methods}  & VOC   & MS~COCO  \\ 
        \midrule
        \multirow{3}*{\rotatebox{90}{\emph{w/o} FC2}}& $\mathcal{L}_{bce}$ only &  48.8 & 33.1 \\
        & $\mathcal{L}_{sce}$ only &  44.6 & 27.9 \\
        & $\mathcal{L}_{sce}$ for single only &  49.4 & 33.4 \\
        \midrule
        \multirow{4}*{\rotatebox{90}{\emph{w/} FC2}}& $\mathbf{w} $ (FC1 weights)  &  52.1 & 34.6 (rp.)\\
        & $\mathbf{w}'$ (FC2 weights) &  54.1 & 33.2\\
        & $\mathbf{w}\oplus \mathbf{w}'$ &  52.7 & 33.7\\
        & $\mathbf{w}\otimes \mathbf{w}'$  &  54.8 (rp.) & 34.0\\
 
    \bottomrule
  \end{tabular}}
  \vspace{-1mm}
  \caption{
  The upper block shows the mIoU results (\%) of training a conventional multi-label classification model with different loss functions: BCE, SCE and their mixture (SCE for single-label images and BCE for multi-label images). The lower block shows the results of extracting ReCAM using different weights: the weights of FC Layer-1 or FC Layer-2 or their mixture variants (element-wise addition or multiplication). ``rp.'' denotes the options we used to report the final results (including the mIoUs of mask refinement and semantic segmentation). \textbf{Please note} the results of using other options (e.g., $\mathbf{w}'$ used for VOC) are in the supplementary.}
  \label{table_ablation}
  \vspace{-5mm}
\end{table}
}

%% file: tables/table_plugin.tex
\setlength{\tabcolsep}{1.9mm}{
\renewcommand\arraystretch{1}
\begin{table}
  \centering
  \scalebox{0.85}{
  \begin{tabular}{llcccc}
    \toprule
    &\multirow{3}*{Methods}& \multicolumn{2}{c}{CAM} & \multicolumn{2}{c}{ReCAM \small{(ours)}} \\
    \cmidrule(r){3-4}\cmidrule(r){5-6}
    && \texttt{mIoU} & \texttt{Time} & \texttt{mIoU} & \texttt{Time} \\
    && \texttt{(\%)} & \texttt{(ut)} & \texttt{(\%)} & \texttt{(ut)} \\
    \midrule
    \multirow{4}*{\rotatebox{90}{VOC}} &ResNet-50~\cite{cam}    & 48.8  & 1.0   & 54.8  & 1.9    \\
    ~&IRN~\cite{irn}                                            & 66.3  & 8.2   & \underline{70.9}  & 9.1    \\
    ~&AdvCAM~\cite{anti}                                        & 55.6  & 316.3 & 56.6  & 317.2   \\
    ~&AdvCAM + IRN                                              & 69.9  & 323.3 & 70.5  & 324.2   \\
    \midrule
    \multirow{4}*{\rotatebox{90}{MS~COCO}} & ResNet-50~\cite{cam}  & 33.1$^*$  & 1.0     & 34.6  & 2.1    \\
    ~&IRN~\cite{irn}                                            & 42.4$^*$  & 8.5    & 44.1  & 9.6    \\
    ~&AdvCAM~\cite{anti}                                        & 35.8$^*$  & 302.5   & 37.8  & 303.8   \\
    ~&AdvCAM + IRN                                              & 45.6$^*$  & 311.0   & \underline{46.3}  & 312.2   \\
    \bottomrule
  \end{tabular}}
  \caption{Comparing ReCAM with baselines in terms of pseudo mask mIoU (\%) and consumption time on VOC and MS~COCO dataset. ``Time'' means the total computing time \underline{from} training the model (with an ImageNet pre-trained backbone) \underline{to} generating 0-1 masks of all training images. The unit time (\texttt{ut}) is 0.7 hours on VOC~\cite{voc} and 5.4 hours for MS~COCO~\cite{mscoco}. $^*$ denotes results are from our re-implementation (no MS~COCO results in original papers). Underline highlights our best results.}
  \label{table_plugin}
  \vspace{-0.4cm}
\end{table}
}

%% file: tables/table_seg.tex
\setlength{\tabcolsep}{1.2mm}{
\renewcommand\arraystretch{1}
\begin{table*}
  \centering
  \begin{tabular}{lcccccccccccc}
    \toprule
    \multirow{4}*{Methods}& \multicolumn{8}{c}{VOC}& \multicolumn{4}{c}{MS~COCO} \\
    \cmidrule(r){2-9}\cmidrule(r){10-13}
    & \multicolumn{4}{c}{DeepLabV2} &\multicolumn{4}{c}{UperNet-Swin}& \multicolumn{2}{c}{DeepLabV2} &\multicolumn{2}{c}{UperNet-Swin}\\
    \cmidrule(r){2-5}\cmidrule(r){6-9}\cmidrule(r){10-11}\cmidrule(r){12-13}
    & \multicolumn{2}{c}{CAM} & \multicolumn{2}{c}{ReCAM}& \multicolumn{2}{c}{CAM} & \multicolumn{2}{c}{ReCAM}& CAM &ReCAM & CAM &ReCAM \\
    \cmidrule(r){2-3}\cmidrule(r){4-5}\cmidrule(r){6-7}\cmidrule(r){8-9}\cmidrule(r){10-10}\cmidrule(r){11-11}\cmidrule(r){12-12}\cmidrule(r){13-13}
    & \texttt{val} & \texttt{test} & \texttt{val} & \texttt{test}& \texttt{val} & \texttt{test} & \texttt{val} & \texttt{test} & \texttt{val} & \texttt{val} & \texttt{val} & \texttt{val} \\
    \hline
    ResNet-50~\cite{cam}    & 54.3  & 55.0  & 59.0\scriptsize{{+4.7}}               & 58.7\scriptsize{+3.7}                 & 48.5  & 49.6  & 54.6\scriptsize{{+6.1}}               & 55.3\scriptsize{{+5.7}}               & 35.7  & 36.5\scriptsize{+0.8}             & 35.9  & 36.8\scriptsize{+0.9}                \\
    IRN~\cite{irn}          & 63.5  & 64.8  & \underline{68.7}\scriptsize{{+5.2}}   & \underline{68.5}\scriptsize{{+3.7}}   & 65.9  & 67.7  & \underline{70.9}\scriptsize{{+5.0}}   &  71.5\scriptsize{{+3.8}}              & 42.0  & 42.9\scriptsize{+0.9}             & 44.0  & 46.0\scriptsize{{+2.0}}              \\
    AdvCAM~\cite{anti}      & 58.3  & 57.9  & 59.1\scriptsize{{+0.8}}               & 59.0\scriptsize{{+1.1}}               & 55.8  & 56.2  & 57.3\scriptsize{{+1.5}}               & 57.4\scriptsize{{+1.2}}               & 37.0  & 39.4\scriptsize{+2.4}             & 37.8  & 39.6\scriptsize{+1.8}                \\
    AdvCAM + IRN            & 68.1  & 68.0  & 68.4\scriptsize{+0.3}                 & 68.2\scriptsize{{+0.2}}               & 70.2  & 70.4  &70.4\scriptsize{{+0.2}}                & \underline{71.7}\scriptsize{{+1.3}}   & 44.2  & \underline{45.0}\scriptsize{+0.8} & 46.8  & \underline{47.9}\scriptsize{+1.1}    \\
    \bottomrule
  \end{tabular}
  \vspace{-2mm}
  \caption{The mIoU results (\%) of WSSS using different segmentation models on two benchmarks. Seed masks are generated by either CAM or ReCAM, and mask refinement methods are row titles. We provide the results of DeepLabV3+ in the supplementary materials.}
  \label{table_seg}
  \vspace{-0.4cm}
\end{table*}
}

%% file: figures/figure_vis.tex
\begin{figure*}[t]
    \centering
    \footnotesize
    \includegraphics[width=1\linewidth]{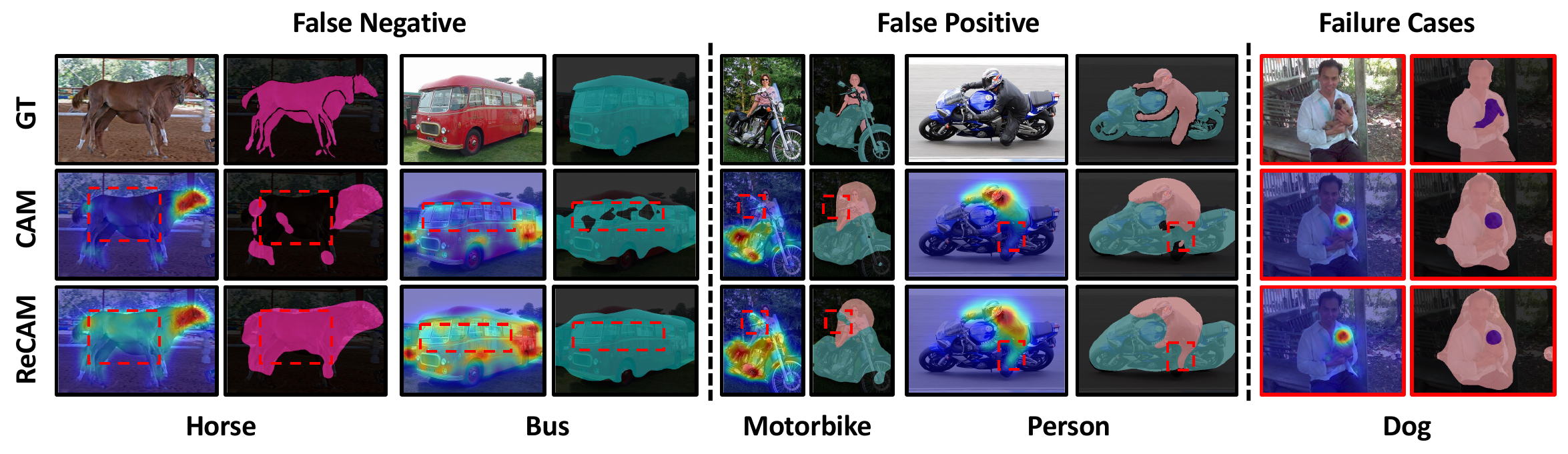}
    \vspace{-3mm}
    \caption{Visualization of 0-1 masks generated by using CAM and ReCAM on the VOC dataset (before training segmentation models). Left two blocks (each with four columns) present the two flaws introduced in Section~\ref{sec_intro}: false negative pixels and false positive pixels, respectively. Red dashed boxes highlight the regions improved by ReCAM. The last block shows an example of failure case.}
    \vspace{-3mm}
    \label{fig:vis}
\end{figure*}

%% file: figures/figure_lambda.tex
\begin{figure}[t]
    \centering
    \footnotesize
    \includegraphics[width=.98\linewidth]{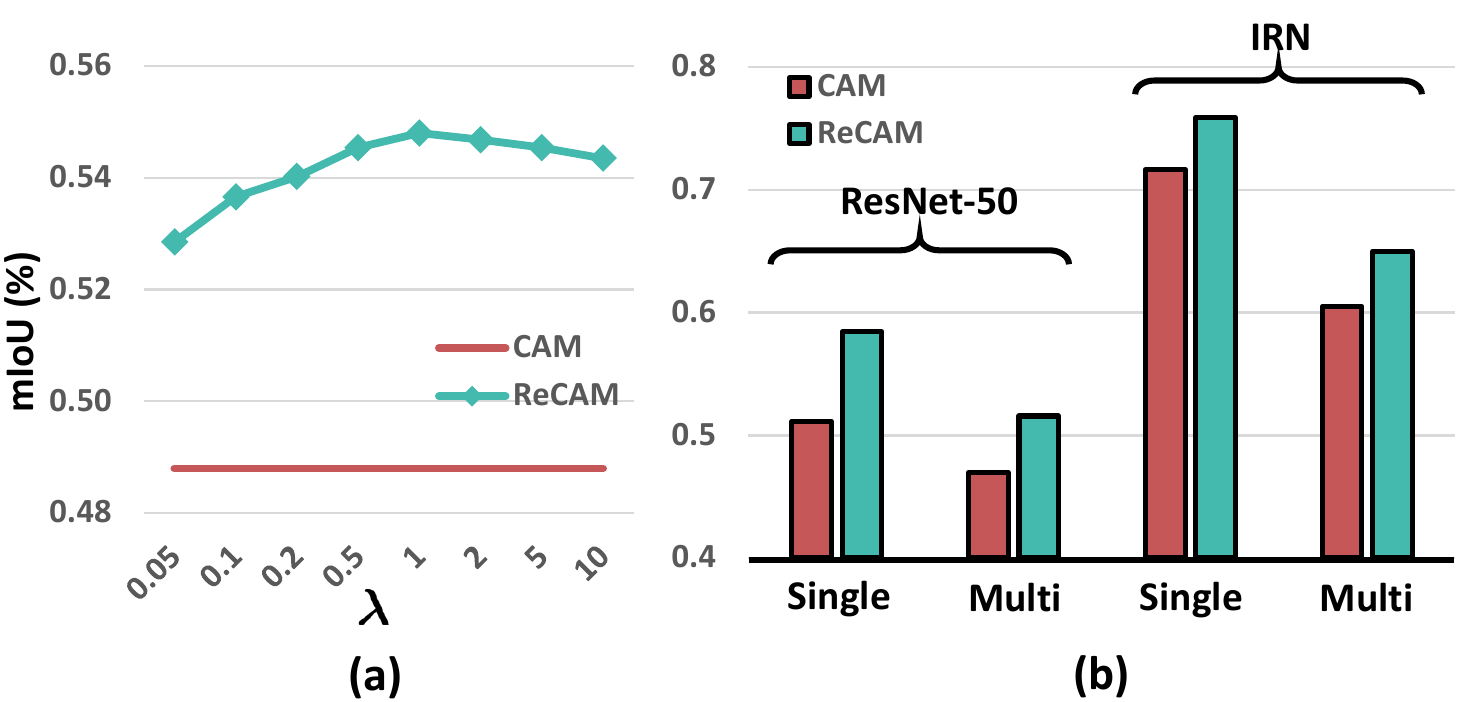}
    \caption{(a) The sensitiveness of ReCAM to the value of $\lambda$ in Eq.~\eqref{eq:loss_ours}, on VOC.
    (b) Decomposing the mIoU results on the first two lines of Table~\ref{table_plugin} into the respective results of single-label images (``Single'') and multi-label (``Multi'') images. }
    \vspace{-3mm}
    \label{fig:lambda}
\end{figure}

%% file: sections/4_experiment_part2.tex
\noindent
\textbf{Generality of ReCAM.}
\emph{We take ReCAM as seed}, and evaluate its generality by: 1) comparing it to the vanilla CAM---the most commonly used seed generation method; and 2) applying different refinement methods after it. From the results in Table~\ref{table_plugin} and~\ref{table_seg}, we can find that ReCAM shows consistent advantages over CAM on both VOC and MS~COCO. Specifically on the first row of Table~\ref{table_plugin}, ReCAM itself outperforms CAM by $6\%$ on VOC. This margin is almost maintained when using ReCAM as pseudo masks to learn semantic segmentation models, as shown in the first row of Table~\ref{table_seg}.
It is worth mentioning that the margin is larger on the stronger segmentation model UperNet-Swin, e.g., $6.1\%$ compared to the $4.7\%$ using DeepLabV2, on VOC \texttt{val}.

For refining ReCAM, we have two observations: 1) the computational cost increases significantly (Table~\ref{table_plugin}), e.g., about $4.5$ times caused by IRN and $160$ times by AdvCAM (over the vanilla ReCAM on ResNet-50); and 2) the best performance of WSSS is achieved always with the help of IRN, as shown in the underlined numbers in Table~\ref{table_seg}.

Figure~\ref{fig:lambda}~(b) shows that ReCAM generates better masks for single-label as well as multi-label images\footnote{Single-label images have the main issues of \underline{false negative pixels} and multi-label images have more \underline{false positive pixels} due to the co-occurring classes. We provide detailed statistics in the supplementary materials.}. The improvements of ReCAM are maintained when adding IRN. Figure~\ref{fig:vis} shows $4$ examples where ReCAM mitigates the two flaws we mentioned in Section~\ref{sec_intro}: \underline{false negative pixels} and \underline{false positive pixels}. The rightmost block in Figure~\ref{fig:vis} shows
a failure case: both CAM and ReCAM fail to capture the object parts with the occlusion or similar color to the surrounding, e.g., between ``dog'' and ``human hands''.

\input{tables/table_sota}

\noindent
\textbf{Superiority of ReCAM.}
\emph{We may also take ReCAM as a refinement method}, and compare it with related methods such as IRN and AdvCAM.
In Table~\ref{table_plugin}, compared to AdvCAM ($55.6\%$), ReCAM achieves a comparable result of $54.8\%$ on VOC, yet it is more efficient---160$\times$ faster than AdvCAM ($1.9$ \texttt{ut} \emph{v.s.} $316.3$ \texttt{ut}). By cascading IRN in addition, ReCAM surpasses AdvCAM by $1\%$ ($70.9\%$ \emph{v.s.} $69.9\%$), and ReCAM is more efficient (only $8.2$ \texttt{ut}). Besides, we can see from Table~\ref{table_sota} that ReCAM supports plug-and-play in different CAM variants including saliency-based methods.

%% file: tables/table_sota.tex
\setlength{\tabcolsep}{1.1mm}{
\renewcommand\arraystretch{1}
\begin{table}
  \centering
    \scalebox{0.85}{
  \begin{tabular}{p{2.3cm}cc|p{2.3cm}cc}
    \toprule
    \emph{w/o} saliency & \texttt{val} & \texttt{test} & \emph{w/} saliency & \texttt{val} & \texttt{test} \\
    \hline
    IRN~\cite{irn}              & 63.5  & 64.8  & DSRG~\cite{dsrg}      & 61.4  & 63.2  \\
    OOA~\cite{ooa}              & 63.9  & 65.6  & OOA*~\cite{ooa}       & 65.2  & 66.4  \\
    ICD~\cite{icd}              & 64.1  & 64.3  & SGAN~\cite{sgan}      & 66.2  & 66.9  \\
    SEAM~\cite{seam}            & 64.5  & 65.7  & ICD~\cite{icd}        & 67.8  & 68.0  \\
    SC-CAM~\cite{sub_category}  & 66.1  & 65.9  & NSROM~\cite{nsrom}    & 68.3  & 68.5  \\
    BES~\cite{bes}              & 65.7  & 66.6  & EDAM*~\cite{edam}     & 70.9  & 70.6  \\
    CONTA~\cite{conta}          & 65.3  & 66.1  & EPS*~\cite{eps}       & 70.9  & 70.8  \\
    \cline{4-6}
    AdvCAM~\cite{anti}          & 68.1  & 68.0 &\multicolumn{3}{l}{\emph{plugin results:}}\\
    RIB~\cite{rib}              & 68.3  & \underline{68.6}  & ReCAM-E*              & 71.6  & 71.4  \\
    ReCAM                       & \underline{68.5}  & 68.4  & ReCAM-M*              & \underline{71.8}  & \underline{72.2}  \\
    \bottomrule
  \end{tabular}}
  \caption{The mIoU results (\%) using DeepLabV2 on VOC, with or without saliency detection models. On the left, the methods are with IRN (by default) if they reported such combo in their papers. On the right, we plug ReCAM respectively into EPS* (-E*) and EDAM* (-M*), or equivalently, adding their saliency encoding modules respectively into our framework, where * denotes DeepLabV2 is pre-trained on MS~COCO.
  }
  \label{table_sota}
  \vspace{-5mm}
\end{table}
}

%% file: sections/5_conclusion.tex
\section{Conclusions}
We started from the two common flaws of the conventional CAM. We pointed out the crux is the widely used BCE loss and demonstrated the superiority of SCE loss theoretically and empirically. We proposed a simple yet effective method named ReCAM by plugging SCE into the BCE-based model to reactivate the model. We showed its generality and superiority via extensive experiments and various case studies on two popular WSSS benchmarks.

\section*{Acknowledgments}

This research was supported by A*STAR under its AME YIRG Grant (Project No. A20E6c0101), and Alibaba Innovative Research (AIR) programme.

%% file: sections/6_supp.tex
\beginsupp

\noindent
{\Large {\textbf{Supplementary materials}}}
\\

We present more details about the toy experiments in Section~\ref{sec:toy},
the quantitative results in Section~\ref{sec:deeplabv3} supplementing for Table~\ref{table_seg} (main paper), 
more quantitative results in Section~\ref{sec:weight_recam} related to Table~\ref{table_ablation} (main paper), 
the statistics of false positive and false negative pixels in Section~\ref{sec:two_flaws},
the pseudo mask quality (mIoU) of ReCAM by traversing the value of $\lambda$ on MS~COCO in Section~\ref{sec:lambda},
the sensitivity analysis to learning rates in Section~\ref{sec:lr},
the detailed derivation of SCE and BCE in Section~\ref{sec:gradient},
the algorithm of ReCAM in Section~\ref{sec:algorithm},
more training details in Section~\ref{sec:train_details},
and more qualitative results in Section~\ref{sec:qualitative} supplementing for Figure~\ref{fig:vis} (main paper).

\section{More Details about Toy Experiments}
\label{sec:toy}
The 5 hoofed-animal classes include \texttt{horse}, \texttt{sheep}, \texttt{cow}, \texttt{elephant}, and \texttt{bear} on the MS~COCO. We chose the images that contain only one of these classes and ignored other MS~COCO classes co-occurring in the image (e.g., the image containing a \texttt{person} and a \texttt{horse} will be selected but it is labeled with a one-hot label \texttt{horse}). There are 6,340 and 3,001 such images in MS~COCO \texttt{train} and \texttt{val} set, respectively. Then, we trained the 5-class classification models on the 6,430 \texttt{train} images using BCE or SCE losses, respectively, and evaluated the models on the 3,001 \texttt{val} images (please note we also showed the class activation results (mIoU) for training images in the main paper).

\section{More WSSS Results (DeepLabV3+)}
\label{sec:deeplabv3}
Table~\ref{table:deeplabv3} presents the mIoU (\%) results of WSSS when exploiting DeepLabV3+ models. It is to supplement for Table~\ref{table_seg} in the main paper.
\input{tables/deeplabv3}

\section{Different Weights for ReCAM}
\input{tables/diff_weights}
Table~\ref{table:diff_weights} shows the mIoU results (\%) of WSSS (using DeepLabV2) when applying different FC weights to extract ReCAM. We show two blocks of WSSS results: one for using ReCAM to generate seeds (and directly using seed masks to train WSSS models), and the other one with a further step of refining the seed masks using IRN (and then using refined masks to train WSSS models).

\section{Statistics of Two Flaws}
\label{sec:two_flaws}
\input{tables/fp_fn}
In Table~\ref{table:fp_fn}, we show the detailed statistics of two flaws analyzed in the main paper. We also provide the numbers of TP and FP (bg) as additional reference. It is worth mentioning that when thresholding CAM to be 0-1 mask, there is a trade-off between FP and FN---a higher threshold value results in less FP and more FN. \emph{Please note that we follow AdvCAM~\cite{anti} to do the fine-grained grid search for the value of the threshold.}

Table~\ref{table:fp_fn} presents the results of using two different thresholds for VOC (threshold=0.21 was used in the main paper). We can see that when comparing ReCAM (0.21) with CAM, there is a significant decrease in FN pixels, but a slight increase in FP pixels (including both obj and bg). When increasing the threshold to 0.26, both FN and FP pixels are reduced. However, the overall improvements over CAM drops a bit (threshold=0.21 results in 54.8\% mIoU, however, threshold=0.26 results in 53.8\% mIoU). We are happy to see that ReCAM reduces both FP and FN clearly when it achieves the best performance on the more challenging dataset---MS~COCO.

\section{$\lambda$ on MS~COCO}
\label{sec:lambda}

\begin{figure}[ht]
\begin{center}
\includegraphics[width=0.8\linewidth]{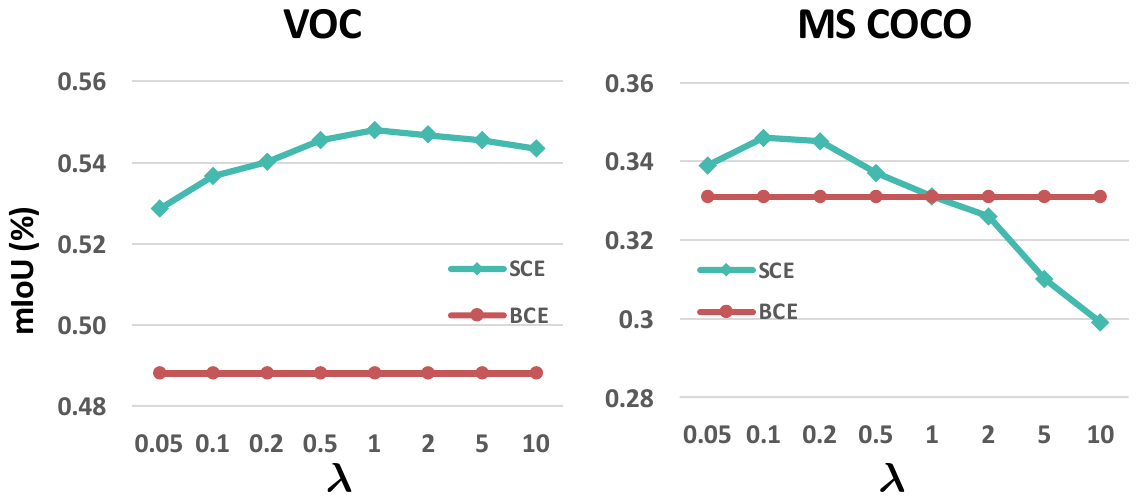}
\end{center}
  \caption{The sensitiveness of ReCAM to the value of $\lambda$ on MS~COCO.}
\label{fig:lambda_coco}
\end{figure}

The hyperparameter $\lambda$ balances the effects of BCE and SCE terms in the loss function (see Eq.~\eqref{eq:loss_ours} in the main paper). We show the effects on the results when traversing the value of $\lambda$ on MS~COCO in Figure~\ref{fig:lambda_coco}. This is to supplement for Figure~\ref{fig:lambda}~(a) of the main paper. The optimal value of $\lambda$ on MS~COCO is 0.1, and the model performance drops a lot when using large $\lambda$ values (e.g. 2). We have explained the reason in the paragraph of Section~\ref{sec_result} entitled as ``Using the Weights of FC1 and FC2 in Eq.(9)''.

\section{Sensitivity to learning rate}
\label{sec:lr}

\begin{figure}[ht]
\begin{center}
    \includegraphics[width=0.8\linewidth]{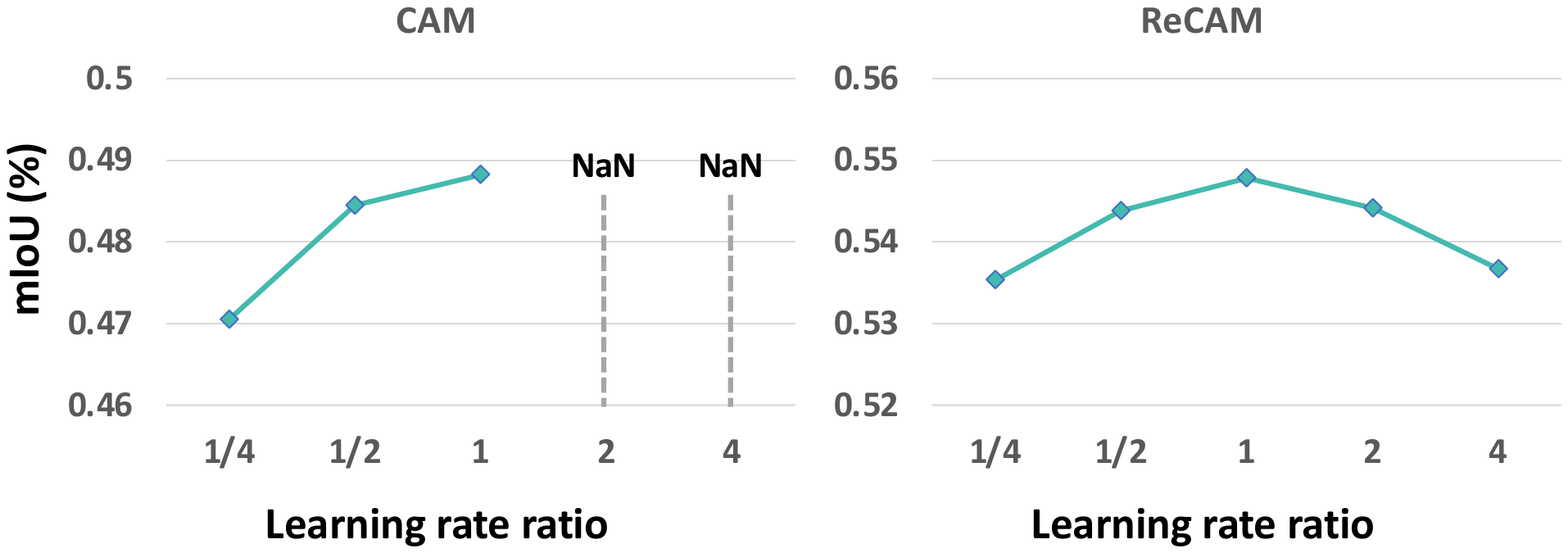}
\end{center}
\caption{The mIoU results of CAM and ReCAM trained with different learning rates on VOC. Learning rate ratio denotes the scalar applied on the default learning rates used in the paper.}
\label{fig:lr}
\end{figure}

We run experiments using different learning rates (LR). We realize this by applying different scalars (denoted as learning rate ratios in Figure~\ref{fig:lr}) on the default LR values used in the paper. Please note that our setting of LR for baseline CAM exactly follows IRN~\cite{irn}. We can see in Figure~\ref{fig:lr} that large LR values make the training of CAM unstable and end with a NaN loss. In contrast, our ReCAM is not sensitive. We think this is because the two FC layers in ReCAM are trained not from scratch but from the weights of a pre-trained baseline BCE model (that is why we call it ``Reactivating CAM''). We have highlighted this reason in the paragraph of Section~\ref{sec_result} entitled as ``Using the Weights of FC1 and FC2 in Eq.(9)''.

\section{Gradients of BCE and CE}
\label{sec:gradient}
The gradients of BCE and SCE losses on logits can be derived as:
\begin{equation}
\setcounter{equation}{14}
\begin{aligned}
    & \nabla_{\bm{z}} \mathcal{L}_{bce} = \frac{\sigma_{sig}(\bm{z})-\bm{y}}{K}, \\
    & \nabla_{\bm{z}} \mathcal{L}_{sce} = \sigma_{sof}(\bm{z})-\bm{y},
\end{aligned}
\end{equation}
here we will show how to compute them.

\subsection{BCE}
BCE Loss in multi-label classification:
\begin{equation}
\mathcal{L}_{bce} = -\frac{1}{K} \sum_i \bm{y}_i \log \sigma_{sig}(\bm{z}_i) + (1-\bm{y}_i) \log (1-\sigma_{sig}(\bm{z}_i)).
\end{equation}

The derivative of sigmoid function is
\begin{equation}
\begin{aligned}
\nabla_{i} \sigma_{sig}(i) &= \nabla_{i} \frac{1}{1+e^{-i}} \\
    &= \frac{e^{-i}}{(1+e^{-i})^2} \\
    &= \sigma_{sig}(i) (1-\sigma_{sig}(i)).
\end{aligned}
\end{equation}

For positive class $p$, $\bm{y}_p$=1, the gradient of $\mathcal{L}_{bce}$ with respect to class $p$ is
\begin{equation}
\begin{aligned}
    \nabla_{\bm{z}_p} \mathcal{L}_{bce} &= -\frac{1}{K} \nabla_{\bm{z}_p} (\bm{y}_p \log \sigma_{sig}(\bm{z}_p))\\
    &= -\frac{1}{K} \frac{\sigma_{sig}(\bm{z}_p) (1-\sigma_{sig}(\bm{z}_p))}{\sigma_{sig}(\bm{z}_p)}\\
    &= \frac{\sigma_{sig}(\bm{z}_p)-1}{K} \\
    &= \frac{\sigma_{sig}(\bm{z}_p)-\bm{y}_p}{K}.
\end{aligned}
\end{equation}

For  negative class $q$, $\bm{y}_q$=0, the gradient of $\mathcal{L}_{bce}$ with respect to class $q$ is
\begin{equation}
\begin{aligned}
    \nabla_{\bm{z}_q} \mathcal{L}_{bce} &= -\frac{1}{K} \nabla_{\bm{z}_q} ((1-\bm{y}_p) \log (1-\sigma_{sig}(\bm{z}_q)))\\
    &= \frac{1}{K} \frac{(1-\sigma_{sig}(\bm{z}_q)) \sigma_{sig}(\bm{z}_q)}{1-\sigma_{sig}(\bm{z}_q)}\\
    &= \frac{\sigma_{sig}(\bm{z}_q)}{K} \\
    &= \frac{\sigma_{sig}(\bm{z}_q)-\bm{y}_q}{K}.
\end{aligned}
\end{equation}

Then, the gradients of $\mathcal{L}_{bce}$ on logits can be derived as
\begin{equation}
\nabla_{\bm{z}} \mathcal{L}_{bce} = \frac{\sigma_{sig}(\bm{z})-\bm{y}}{K}.
\end{equation}

\subsection{CE}

SCE Loss in multi-label classification:
\begin{equation}
\mathcal{L}_{sce} = - \sum_i \bm{y}_i \log \sigma_{sof}(\bm{z}_i).
\end{equation}

For positive class $p$, $\bm{y}_p$=1, the gradient of $\mathcal{L}_{sce}$ with respect to class $p$ is
\begin{equation}
\begin{aligned}
    \nabla_{\bm{z}_p} \mathcal{L}_{sce} & = \nabla_{\bm{z}_p} (-\log e^{\bm{z}_p} + \log \sum_i e^{\bm{z}_i}) \\
    & = -1+\frac{e^{\bm{z}_p}}{\sum_i e^{\bm{z}_i}} \\
    & = \sigma_{sof}(\bm{z}_p)-\bm{y_p}.
\end{aligned}
\end{equation}

For  negative class $q$, $\bm{y}_q$=0, the gradient of $\mathcal{L}_{sce}$ with respect to class $q$ is
\begin{equation}
\begin{aligned}
    \nabla_{\bm{z}_q} \mathcal{L}_{sce} & = \nabla_{\bm{z}_q} (-\log e^{\bm{z}_p} + \log \sum_i e^{\bm{z}_i}) \\
    & = \frac{e^{\bm{z}_q}}{\sum_i e^{\bm{z}_i}} \\
    & = \sigma_{sof}(\bm{z}_q)-\bm{y}_q.
\end{aligned}
\end{equation}

Then, the gradients of $\mathcal{L}_{sce}$ on logits can be derived as
\begin{equation}
\nabla_{\bm{z}} \mathcal{L}_{sce} = \sigma_{sof}(\bm{z})-\bm{y},
\end{equation}

\section{Algorithm}
The training pipeline of ReCAM is in Algorithm~\ref{alg}.
\label{sec:algorithm}

\begin{algorithm}
\caption{ReCAM}\label{alg}
\renewcommand{\algorithmicrequire}{\textbf{Input:}}
\renewcommand{\algorithmicensure}{\textbf{Output:}}
\begin{algorithmic}[1]
\Require Training images $\mathcal{X}$ and image-level labels $\mathcal{Y}$
\Ensure ReCAM (soft masks before thresholding)

\State Load ImageNet pre-trained feature extractor $\Theta$
\State Randomly initialize weights $\mathbf{w}$ of FC Layer-1
\For{Image batches in $\mathcal{X}$}
    \State Optimize $\Theta$ and $\mathbf{w}$ using Eq. (1)
\EndFor
\State Randomly initialize weights $\mathbf{w}$' of FC Layer-2
\For{Image batches in $\mathcal{X}$}
    \State Optimize $\Theta$, $\mathbf{w}$, and $\mathbf{w}'$ using Eq. (8)
\EndFor
\State Extract ReCAM using Eq. (9)
\end{algorithmic}
\end{algorithm}

\section{Training Details of DeepLabV2}
\label{sec:train_details}
We supplement the details in the training process of DeepLabV2. Following~\cite{anti,rib}, we cropped each training image to the size of 321$\times$321. We trained the model for 20k and 100k iterations on VOC and MS~COCO datasets, respectively, with the respective batch size of 5 and 10. The learning rate was set as 2.5e-4 and weight decay as 5e-4. Horizontal flipping and random crop were used for data augmentation. 

\section{More Qualitative Results}
\label{sec:qualitative}
\begin{figure*}[ht]
\begin{center}
\includegraphics[width=0.98\linewidth]{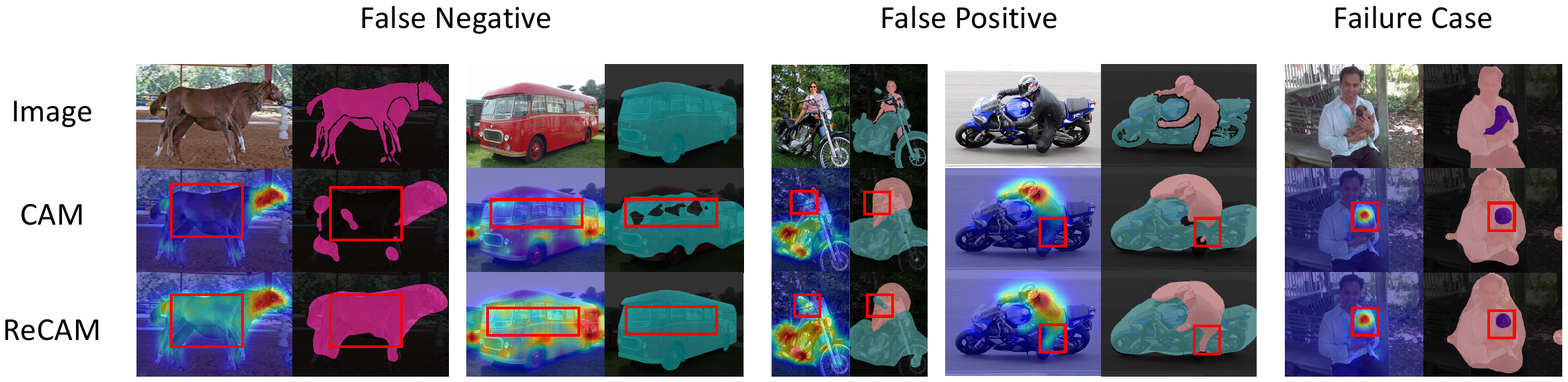}
\end{center}
  \caption{Visualization of the soft masks and 0-1 masks, generated by CAM and ReCAM on the VOC dataset.}
\label{fig:vis_seed}
\end{figure*}

\begin{figure*}[ht]
\begin{center}
\includegraphics[width=0.98\linewidth]{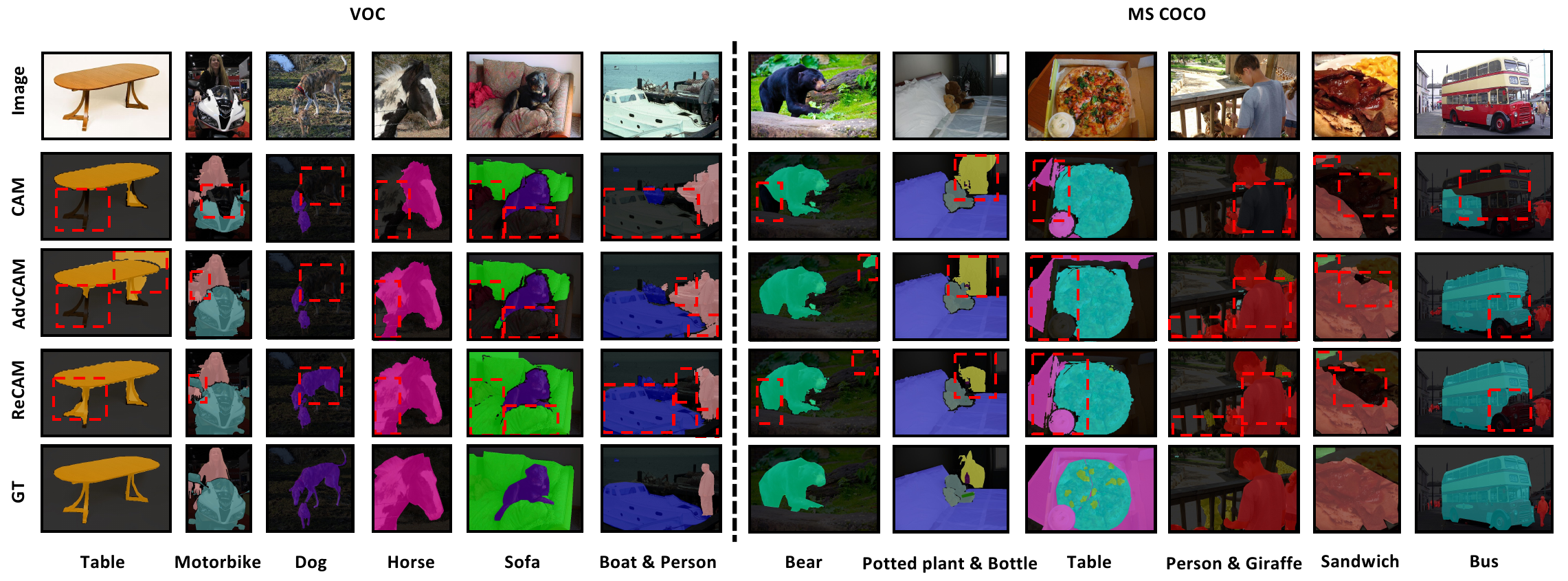}
\end{center}
  \caption{Visualization of the masks (from different CAM variants) refined by IRN~\cite{irn} on the VOC and MS~COCO datasets.}
\label{fig:vis_pseudo}
\end{figure*}

\begin{figure*}[ht]
\begin{center}
\includegraphics[width=0.98\linewidth]{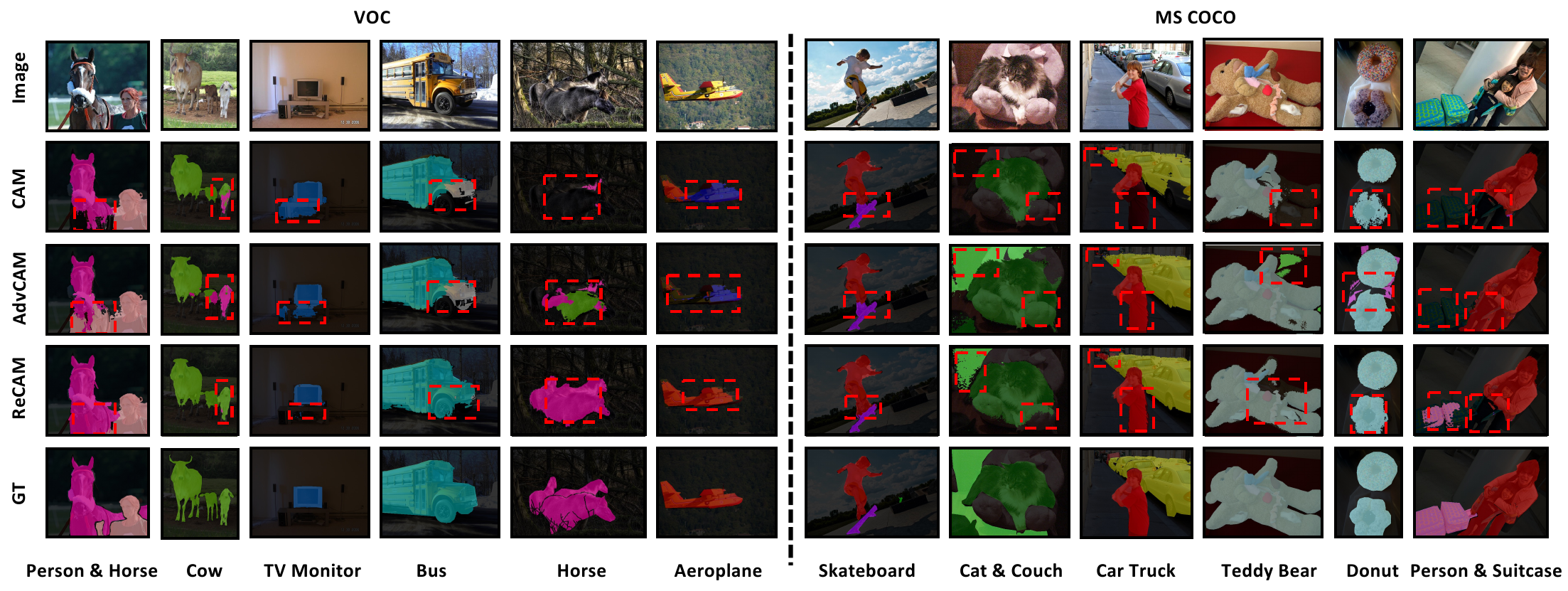}
\end{center}
  \caption{Visualization of the semantic segmentation results (i.e., predicted masks) using DeepLabV2 on the VOC and MS~COCO datasets. The seeds (of pseudo masks) are generated by different methods first and then are all refined by IRN.}
\label{fig:vis_ss}
\end{figure*}

%
Figure~\ref{fig:vis_seed} shows more qualitative results of heatmaps and 0-1 masks generated by CAM and ReCAM, on VOC \texttt{train} set. This is to supplement for Figure~\ref{fig:vis} in the main paper.
Figure~\ref{fig:vis_pseudo} shows the refined masks using IRN~\cite{irn} on VOC and MS COCO \texttt{train} set.
Figure~\ref{fig:vis_ss} shows the resulted masks of semantic segmentation (using DeepLabV2) on VOC and MS COCO \texttt{val} set.

%% file: tables/deeplabv3.tex
\setlength{\tabcolsep}{1.2mm}{
\renewcommand\arraystretch{1}
\begin{table}[ht]
  \centering
  \begin{tabular}{lcccc}
    \toprule
    \multirow{2}*{Methods}& \multicolumn{2}{c}{VOC}& \multicolumn{2}{c}{MS~COCO} \\
    \cmidrule(r){2-3}\cmidrule(r){4-5}
    & CAM &ReCAM& CAM &ReCAM \\
    \hline
    ResNet-50~\cite{cam}    & 53.5  & 57.8  & 36.2  & 37.0  \\
    IRN~\cite{irn}          & 64.2  & 69.3  & 43.1  & 44.3  \\
    AdvCAM~\cite{anti}      & 57.3  & 58.3  & 37.7  & 40.2  \\
    AdvCAM + IRN            & 68.5  & 68.6  & 44.4  & 45.8  \\
    \bottomrule
  \end{tabular}
  \vspace{3mm}
  \caption{The semantic segmentation results (mIoU\%) of using DeepLabV3+, on VOC and MS~COCO. Seed masks are generated by either CAM or ReCAM, and then fed into mask refinement algorithms (listed as row titles). The refined masks are used as pseudo labels to train the semantic segmentation model. Finally, the model is evaluated on the \texttt{val} set.}
  \label{table:deeplabv3}
\end{table}
}

%% file: tables/diff_weights.tex
\label{sec:weight_recam}
\setlength{\tabcolsep}{1.8mm}{
\renewcommand\arraystretch{1}
\begin{table}[ht]
  \centering
  \scalebox{1}{
  \begin{tabular}{llcc}
    \toprule
    &Weights& VOC & MS~COCO \\
    \midrule
    \multirow{3}*{ResNet-50~\cite{resnet}}&$\mathbf{w}$ (FC1 weights)       & 56.5              & \underline{36.5}  \\
    ~&$\mathbf{w}'$ (FC2 weights)                                           & 58.2              & 35.6              \\
    ~&$\mathbf{w}\otimes \mathbf{w}'$                                       & \underline{59.0}  & 36.1              \\
    \midrule
    \multirow{3}*{IRN~\cite{irn}} & $\mathbf{w}$ (FC1 weights)              & 64.8              & \underline{42.9}  \\
    ~&$\mathbf{w}'$ (FC2 weights)                                           & 67.5              & 42.1              \\
    ~&$\mathbf{w}\otimes \mathbf{w}'$                                       & \underline{68.7}  & 42.6              \\
    \bottomrule
  \end{tabular}}
  \vspace{0.4cm}
  \caption{The semantic segmentation results (mIoU\%) using DeepLabV2 on VOC and MS~COCO. It is to show the difference of using different FC weights for computing ReCAM. The pseudo masks (used to train the models) are either ReCAM in the first block or refined masks of ReCAM with IRN in the second block.}
  \label{table:diff_weights}
\end{table}
}

%% file: tables/fp_fn.tex
\setlength{\tabcolsep}{1.0mm}{
\renewcommand\arraystretch{1}
\begin{table}[ht]
  \centering
  \scalebox{0.65}{
  \begin{tabular}{lccccc}
    \toprule
    ~&\multicolumn{3}{c}{VOC}  &\multicolumn{2}{c}{MS~COCO}  \\ 
    \cmidrule(r){2-4}\cmidrule(r){5-6}
    ~& \multicolumn{1}{c}{CAM} & \multicolumn{1}{c}{ReCAM (0.21)} & \multicolumn{1}{c}{ReCAM (0.26)} & \multicolumn{1}{c}{CAM} & \multicolumn{1}{c}{ReCAM} \\
    \midrule
    TP      & 2.05e8 (82.5)  & 2.09e8 (84.3)  & 2.1e8 (84.9)   & 1.69e10 (74.1) & 1.72e10 (75.3)  \\
    \hline
    FP (obj) & 1.86e6 (0.8)  & 2.08e6 (0.8)   & 1.80e6 (0.7)   & 7.94e8 (3.5)   & 7.45e8 (3.3)   \\
    FN      & 2.33e7 (9.4)   & 1.65e7 (6.7)   & 2.09e7 (8.4)   & 2.76e9 (12.1)  & 2.60e9 (11.4)   \\
    \hline
    FP (bg)  & 2.02e7 (8.1)  & 2.25e7 (9.1)   & 1.66e7 (6.7)   & 3.16e9 (13.9)  & 3.03e9 (13.3)   \\

    \bottomrule
  \end{tabular}}
  \vspace{0.4cm}
  \caption{The number of pixels and the percentage (\%) for different pixels in the seed masks. ``TP'' indicates the true positive. ``FP (obj)'' indicates the false positive whose actual label is another object class. ``FP (bg)'' is false positive whose actual label is \texttt{background}. ``FN'' is false negative which is misclassified as \texttt{background}.
  The summation of the percentages of ``TP'', ``FN'' and ``FP (bg)'' is 100\% in each column.}
  \label{table:fp_fn}
\end{table}
}